\documentclass{article}

\PassOptionsToPackage{numbers, compress}{natbib}

\usepackage[preprint]{neurips_2024}

\usepackage[utf8]{inputenc} 
\usepackage[T1]{fontenc}    
\usepackage[colorlinks=true, linkcolor=blue, urlcolor=blue, citecolor=blue]{hyperref}       
\usepackage{url}            
\usepackage{booktabs}       
\usepackage{amsfonts}       
\usepackage{nicefrac}       
\usepackage{microtype}      
\usepackage[usenames,dvipsnames]{xcolor}

\usepackage{lipsum}
\usepackage[pdftex]{graphicx}
\usepackage{bbding}
\usepackage{soul}
\usepackage{bibunits}
\usepackage{bm}
\usepackage{natbib}
\usepackage{array}
\usepackage{caption}
\usepackage{subcaption}
\usepackage{pdflscape}
\usepackage{makecell}
\usepackage{framed,multirow}
\usepackage{threeparttable}
\usepackage{amsmath}
\usepackage{amssymb}
\usepackage{pifont}
\usepackage{algorithm}
\usepackage{algpseudocode}
\usepackage{listings}
\usepackage{pythonhighlight}
\usepackage{enumitem}  
\usepackage{listings}
\usepackage{xcolor}
\usepackage{geometry}
\usepackage[many,listings]{tcolorbox}

\definecolor{IceBlue}{HTML}{e0f7fa}
\definecolor{NavyBlue}{HTML}{000080}
\definecolor{darkbrown}{HTML}{8B4513}
\definecolor{pinkcolor}{RGB}{255,228,225}
\definecolor{commentgray}{rgb}{0.5, 0.5, 0.5}
\definecolor{additiongreen}{rgb}{0, 0.6, 0}
\definecolor{deletionred}{rgb}{1, 0, 0}
\definecolor{titlebgdark}{HTML}{CF4F20}
\definecolor{contentbglight}{HTML}{FFF0F0}

\DeclareCaptionFormat{mystyle}{\textbf{#3}}
\newtcolorbox{dialogbox}[1][]{
    colback=IceBlue,
    colframe=NavyBlue,
    fonttitle=\bfseries,
    title=#1,
    listing options={
        basicstyle=\small\ttfamily,
        breaklines=true,
        xleftmargin=0pt,
        xrightmargin=0pt,
        alsoletter={0123456789},
        morekeywords={Content1}, keywordstyle=\color{darkbrown}
    },
    verbatim
}
\newtcolorbox{codebox}[1][]{  
    colback=contentbglight,
    colframe=titlebgdark,
    fonttitle=\bfseries,  
    title=#1,  
    listing options={  
        language=Python,  
        basicstyle=\small\ttfamily,  
        breaklines=true,  
        showspaces=false,
        showtabs=false,
        keepspaces=true,
        tabsize=4,
        columns=flexible,
        xrightmargin=0pt,  
        framexleftmargin=20pt,  
        alsoletter={0123456789},  
        morekeywords={Content1},  
        keywordstyle=\color{darkbrown}  
    },  
}  

\title{PIKE-RAG: sPecIalized KnowledgE and Rationale Augmented Generation}

\author{%
  Jinyu Wang$^\ast$\quad
  Jingjing Fu\thanks{Equal Contribution.} \quad
  Rui Wang\quad Lei Song\quad
  Jiang Bian \\
  \\
  Microsoft Research Asia\\
  \texttt{\{jinywan, jifu, ruiwa, lesong, jiabia\}@microsoft.com} \\
}

\begin{document}

\maketitle

\begin{abstract}

Despite notable advancements in Retrieval-Augmented Generation (RAG) systems that expand large language model (LLM) capabilities through external retrieval, these systems often struggle to meet the complex and diverse needs of real-world industrial applications. The reliance on retrieval alone proves insufficient for extracting deep, domain-specific knowledge performing in logical reasoning from specialized corpora. To address this, we introduce sPecIalized KnowledgE and Rationale Augmentation Generation (PIKE-RAG), focusing on extracting, understanding, and applying specialized knowledge, while constructing coherent rationale to incrementally steer LLMs toward accurate responses. Recognizing the diverse challenges of industrial tasks, we introduce a new paradigm that classifies tasks based on their complexity in knowledge extraction and application, allowing for a systematic evaluation of RAG systems' problem-solving capabilities. This strategic approach offers a roadmap for the phased development and enhancement of RAG systems, tailored to meet the evolving demands of industrial applications. Additionally, we propose knowledge atomizing and knowledge-aware task decomposition to effectively extract multifaceted knowledge from the data chunks and iteratively construct the rationale based on original query and the accumulated knowledge, respectively, showcasing exceptional performance across various benchmarks. Furthermore, we introduce a trainable knowledge-aware decomposer that incorporates domain-specific
rationale into the task decomposition and result-seeking process. The code is publicly available at~\href{https://github.com/microsoft/PIKE-RAG?tab=readme-ov-file}{https://github.com/microsoft/PIKE-RAG}.


\end{abstract}

\section{Introduction}

Large Language Models (LLMs) have revolutionized the field of natural language processing by demonstrating the capability to generate coherent and contextually relevant text. These advanced models are trained on expansive corpora, equipping them with the versatility to execute a diverse spectrum of linguistic tasks, ranging from text completion to translation and summarization~\cite{ achiam2023gpt4, bahrini2023chatgpt, touvron2023llama, anil2023gemini}. 
Despite their broad capabilities, LLMs exhibit pronounced limitations when tasked with specialized queries in professional domains~\cite{ling2024domainspecializationkeymake,wang2023survey}, a demand that is particularly acute in industrial applications. This primarily stems from the scarcity of domain-specific training material and a limited grasp of specialized knowledge and rationale within these domains. As a result, LLMs may produce responses that are not only potentially erroneous but also lack the detail and precision required for expert-level engagement~\cite{bender2021dangers}. 
Besides the limitations in the domain-specific tasks, another striking issue with LLMs is the phenomena known as "hallucination", where the model generates information that is not grounded in reality or factual data~\cite{beltagy2020fact, xu2024hallucinationinevitableinnatelimitation}. 
Moreover, the knowledge base of LLMs, being static and crystallized at the point of their last update, introduces temporal stasis~\cite{brown2020language}. 
Further compounding these challenges is the issue of long-context comprehension~\cite{li2024longcontextllmsstrugglelong}. 
Existing LLMs struggle to maintain an understanding of task definitions across long context, and their performance tends to deteriorate significantly when confronted with more complex and demanding tasks.

To address the inherent limitations of LLMs, Retrieval-Augmented Generation (RAG)~\cite{lewis2020retrieval} has been proposed, which merges the generative capabilities of LLMs with a retrieval mechanism, allowing the incorporation of relevant external information to anchor the generated text in factual data. This integrated strategy improves both the accuracy and reliability of the generated content, providing a promising pathway for the practical deployment of LLMs in industrial applications. However, current RAG methods remain heavily reliant on text retrieval and the comprehension capabilities of LLMs, with a lack of attention to extracting, understanding, and utilizing knowledge from the diverse source data. In industrial applications requiring expertise, such as specialized knowledge and problem-solving rationale, existing RAG approaches primarily designed for research benchmarks demonstrate significant limitations. There is a lack of clarity regarding the challenges that RAG encounters in industrial applications. Gaining a comprehensive insight into these challenges is crucial for the development of RAG algorithms. Therefore, we summarize the main challenges as follows.



\begin{itemize}[left=6pt] 

\item \textbf{Knowledge source diversity}: 
RAG systems are constructed upon a diverse corpus of source documents collected over many years from various domains, encompassing a wide range of file formats like scanned images, digital text files, and web data, sometimes accompanied by specialized databases. 
In contrast, widely-used datasets~\cite{ho2020twowiki, yang2018hotpotqa, trivedi2022musique} typically feature pre-segmented, simplified corpora that do not capture the complexity of real-world data.
Existing methods designed for these benchmarks struggle to efficiently extract specialized knowledge and uncover underlying rationales from diverse sources, particularly in industrial applications. 
\textit{For example,} an LED product datasheet typically comprises specifications such as performance characteristics presented in complex tables, electrical properties depicted in charts, and installation instructions illustrated with figures. Addressing queries related to the non-textual knowledge presents significant challenges for existing RAG approaches.


\item \textbf{Domain specialization deficit}: 
In industrial applications, RAG are expected to leverage the specialized knowledge and rationale in professional fields. However, these specialized knowledge are characterized by domain-specific terminologies, expertise, and distinctive logical frameworks that are integral to their functioning. 
RAG approaches built on common knowledge-centric datasets demonstrate unsatisfactory performance when applied to professional fields, as LLMs exhibit deficiencies in extracting, understanding, and organizing domain specific knowledge and rationale~\cite{ling2024domainspecializationkeymake}.
\textit{For example,} in the field of semiconductor design, research relies heavily on a deep understanding of underlying physical properties. When LLMs are utilized to extract and organize the specialized knowledge and rationale from the research documents, they often fail to properly capture essential physical principles and achieve a comprehensive understanding due to their inherent limitations. Consequently, RAG systems frequently produce incomplete or inaccurate interpretations of critical problem elements and generate responses that lack proper rationale grounded in physical principles.
Moreover, assessing the quality of professional content generation poses a significant challenge. This issue not only impedes the development and optimization of RAG algorithms but also complicates their practical deployment across various industrial applications.

\item \textbf{One-size-fits-all}: 
Various RAG application scenarios, although based on a similar framework, present different challenges that require diverse capabilities, particularly for extracting, understanding, and organizing domain-specific knowledge and rationale. The complexity and focus of questions vary across these scenarios, and within a single scenario, the difficulty can also differ. \textit{For example,} in rule-based query scenarios, such as determining the legal conditions for mailing items, RAG systems primarily focus on retrieving relevant factual rules by bridging the semantic gap between the query and the rules. In multihop query scenarios, such as comparing products across multiple aspects, RAG systems emphasize extracting information from diverse sources and performing multihop reasoning to arrive at accurate answers. 
Most existing RAG approaches~\cite{zhao2024survey} adopt a one-size-fits-all strategy, failing to account for the varying complexities and specific demands both within and across scenarios. This results in solutions that do not meet the comprehensive accuracy standards required for practical applications, thereby limiting the development and integration of RAG systems in real-world environments.

\end{itemize}


We believe that the key to addressing these challenges lies in advancing beyond traditional retrieval augmentation, by effectively extracting, understanding, and applying specialized knowledge, and developing appropriate reasoning logic tailored to the specific tasks and the knowledge involved. We refer to this approach as \textit{sPecIalized Knowledge and Rationale Augmentation}. Given that various tasks require diverse capabilities, particularly for extracting, understanding, and organizing domain-specific knowledge and rationale, we summarize and categorize the questions commonly encountered into four types with respect to their difficulty: factual questions, linkable-reasoning questions, predictive questions, and creative questions.  Accordingly, we propose a classification of RAG system capability levels, aligned with the system's ability to solve these different types of problems. This classification serves as a guideline for systematically advancing the system's capabilities in a controllable and measurable manner. 

Furthermore, we propose s\textbf{P}ec\textbf{I}alized \textbf{K}nowledg\textbf{E} and \textbf{R}ationale \textbf{A}ugmented \textbf{G}eneration (\textbf{PIKE-RAG}) framework, which not only support phased system development and deployment, demonstrating excellent versatility, but also enhances capabilities by effectively leveraging specialized knowledge and rationale.
Within this framework, knowledge extraction components are employed to extract specialized knowledge from diverse source data, laying a robust foundation for knowledge-based retrieval and reasoning. Additionally, a task decomposer is utilized to dynamically manage the routing of retrieval and reasoning operations, creating specialized rationale based on available knowledge. 
PIKE-RAG enables a phased exploration of RAG capabilities, which facilitates the progressive refinement of RAG algorithms and the staged implementation of RAG applications.
For each developing phase, the RAG framework and its modules are tailored to address specific challenges. For example, in the knowledge base construction phase, a multi-layer heterogeneous graph is employed to effectively represent relationship between
various components of the data, enhancing knowledge organization and integration. The RAG system, designed for factual questions, introduces multi-granularity retrieval, allowing for multi-layer, multi-granularity retrieval across a heterogeneous knowledge graph to improve factual retrieval accuracy. In the advanced RAG system, aiming at addressing complex queries, knowledge atomizing is introduced to fully explore the intrinsic knowledge from data chunks, while knowledge-aware task decomposition manages the retrieval and organization of multiple pieces of atomic knowledge to construct a coherent rationale. 

Extensive experiments are conducted to evaluate the performance of the proposed PIKE-RAG framework on both open-domain and legal benchmarks, and experimental results demonstrate the effectiveness of PIKE-RAG. Our framework and staged development strategy could further advance the current research and application of RAG in industrial contexts. 
In summary, the contributions of this work are as follows:

\begin{itemize}

\item We propose that specialized knowledge and rationale should serve as the core foundation for augmentation, empowering the resolution of tasks that current retrieval-augmented frameworks are unable to solve effectively. Therefore, we introduce a new paradigm that classifies tasks to distinct types based on their difficulty in the knowledge extraction, comprehension and utilization, offering a novel conceptual framework for system design and evaluation. By applying this paradigm, RAG systems' capabilities are stratified to support phased development, particularly enhancing their application in industrial settings.

\item We introduce specialized Knowledge and Rationale Augmented Generation (PIKE-RAG) framework, which is primarily designed with a focus on specialized knowledge extraction and rationale construction. PIKE-RAG enhances the system capabilities by effectively extracting, comprehending, and organizing specialized knowledge and rationale. Additionally, it can customize the system framework to meet varying levels of capability requirements, demonstrating exceptional versatility.

\item We propose knowledge atomizing and knowledge-aware task decomposition to tackle the complex questions, such as multihop queries, achieving significant performance improvements on the multihop benchmarks, particularly in scenarios involving more than two hops. This demonstrates that task decomposition effectively breaks down complex questions into atomic questions, enabling efficient retrieval and organization of atomic knowledge and constructing a coherent rationale to arrive at accurate answers.

\item We introduce a knowledge-aware task decomposer training strategy that involves collecting rationale-driven data by sampling context and creating diverse interaction. This approach trains the decomposer to incorporate domain-specific rationale into both task decomposition and result-seeking trajectories.

\end{itemize}

\section{Related work}
\subsection{RAG}  
Retrieval-Augmented Generation (RAG) has emerged as a promising solution that effectively incorporates external knowledge to enhance response generation. Initially, retrieval-augmented techniques were introduced to improve the performance of pre-trained language models on knowledge-intensive tasks~\cite{lewis2020retrieval, izacard2022atlasfewshotlearningretrieval, borgeaud2022improvinglanguagemodelsretrieving}. With the booming of Large Language Models~\cite{achiam2023gpt4, bahrini2023chatgpt, touvron2023llama, anil2023gemini}, most research in the RAG paradigm has shifted towards a framework that initially retrieves pertinent information from external data sources and subsequently integrates it into the context of the query prompt as supplementing knowledge for contextually relevant generation~\cite{ram2023incontext}.
Following this framework, naive RAG research paradigm~\cite{gao2023retrieval} converts raw data into uniform plain text and segment it into smaller chunks, which are encoded into vector space for query-based retrieval. The top k relevant chunks are used to expand the context of the prompt for generation. To enhance the retrieval quality of the naive RAG, advanced RAG approaches implement specific enhancements across the pre-retrieval, retrieval, and post-retrieval processes, including query optimization~\cite{ma2023query, zheng2023take}, multi-granularity chunking~\cite{chen2023densex, zhong2024mixofgranularityoptimizechunkinggranularity}, mixed retrieval and chunk re-ranking. 

Beyond the aforementioned RAG paradigms, numerous sophisticated enhancements in RAG pipelines and system modules are introduced within modular RAG systems~\cite{gao2024modularragtransformingrag}, aiming to improve system capability and versatility. These advancements have enabled the processing of a wider variety of source data, facilitating the transformation of raw information into structured data and, ultimately, into valuable knowledge~\cite{wang2023knowledgegraphpromptingmultidocument,edge2024localglobalgraphrag}. Furthermore, the indexing and retrieval modules have been refined with multi-granularity and multi-architecture approaches~\cite{yang2023advanced,zhong2024mixofgranularityoptimizechunkinggranularity}. Various pre-retrieval~\cite{gao2022precisezeroshotdenseretrieval,zheng2024stepbackevokingreasoning} and post-retrieval~\cite{cohere2023rerank, jiang2023longllmlingua} functions are proposed to enhance both the retrieval effectiveness and the quality of sequential generation. It has been recognized that naïve RAG systems are insufficient to tackle complex tasks such as summarization~\cite{WikiAsp2021} and multi-hop reasoning~\cite{trivedi2022musique,ho2020twowiki}. Consequently, most recent research focuses on developing advanced coordination schemes that leverage existing modules to collaboratively address these challenges. ITERRETGEN~\cite{shao2023enhancingretrievalaugmentedlargelanguage} and DSP~\cite{khattab2023demonstratesearchpredictcomposingretrievallanguage} employ retrieve-read iteration to leverage generation response as the context for next round retrieval. FLARE~\cite{jiang2023activeretrievalaugmentedgeneration} proposes a confidence-based active retrieval mechanism that dynamically adjusts query with respect to the low-confidence tokens in the regenerated sentences. These loop-based RAG pipelines progressively converge towards the correct answer and provide enhanced flexibility to RAG systems in addressing diverse requirements.

  

\subsection{Knowledge bases for RAG}
In naïve RAG approaches, source data is converted to plain text and chunked for retrieval. However, as RAG applications expand and demand for diversity grows, plain text-based retrieval becomes insufficient for several reasons: (1) textual information is generally redundant and noisy, leading to decreased retrieval quality; (2) complex problems require the integration of multiple data sources, and plain text alone cannot adequately represent the intricate relationships between objects. As a result, researchers are exploring diverse data sources to enrich the corpus, incorporating search engines~\cite{yang2024cragcomprehensiverag, vu2023freshllmsrefreshinglargelanguage}, databases~\cite{wang2023knowledgptenhancinglargelanguage, pan2022endtoendtablequestionanswering, roychowdhury2024erattaextremeragtable}, knowledge graphs~\cite{sun2024thinkongraphdeepresponsiblereasoning, wang2023knowledgegraphpromptingmultidocument}, and multimodal corpora~\cite{chen2022muragmultimodalretrievalaugmentedgenerator, caffagni2024wikillavahierarchicalretrievalaugmentedgeneration}. Concurrently, there is an emphasis on developing efficient knowledge representations for corpus to enhance knowledge retrieval. 
A graph is regarded as a powerful knowledge representation because of its capacity to intuitively model complex relationships. GraphRAG \cite{edge2024localglobalgraphrag} combines knowledge graph generation and query-focused summarization with RAG to address both local and global questions. HOLMES~\cite{panda2024holmeshyperrelationalknowledgegraphs} construct hyper-relational KGs and prune them to distilled graphs, which serve as an input to LLMs for multihop question answering. However, the construction of knowledge graphs is extremely resource-intensive, and the associated costs scale up with the size of the corpus.

\subsection{Multi-hop QA}  
Multi-hop Question Answering (MHQA)~\cite{yang2018hotpotqa} involves answering questions that require reasoning over multiple pieces of information, often scattered across different documents or paragraphs. This task presents unique challenges as it necessitates not only retrieving relevant information but also effectively combining and reasoning over the retrieved pieces to arrive at a correct answer.
The traditional graph-based methods in MHQA solve the problem by building graphs and inferring on graph neural networks(GNN) to predict answers~\cite{qiu2019dynamicgraphs, fang2020hierarchicalgraphs}. With the advent of LLMs, recent graph-based methods~\cite{li2023leveragingstructuredinformationexplainable, panda2024holmeshyperrelationalknowledgegraphs} have evolved to construct knowledge graphs for retrieval and generate response through LLMs.  Another branch of methods dynamically convert multi-hop questions into a series of sub-queries by generating subsequent questions based on the answers to previous ones~\cite{trivedi2023interleavingretrievalchainofthoughtreasoning, khattab2023demonstratesearchpredictcomposingretrievallanguage, feng2023retrievalgenerationsynergyaugmentedlarge}. The subqueries guides the sequential retrieval and the retrieved results in turn are used to improve reasoning. Treating MHQA as a supervised problem, Self-RAG~\cite{zhang2024endtoendbeamretrievalmultihop} trains an LM to learn to retrieve, generate,
and critique text passages, and beam-retrieval~\cite{asai2023selfraglearningretrievegenerate} models the multi-hop retrieval process in an end-to-end manner by jointly optimizing an encoder and classification heads across all hops. Self-Ask~\cite{press2023measuringnarrowingcompositionalitygap} improves CoT by explicitly asking itself follow-up questions before answering the initial question. This method enables the automatic decomposition of questions and can be seamlessly integrated with retrieval mechanisms to tackle Multi-hop Question Answering.  



\section{Problem formulation}

Existing research mainly concentrates on algorithmic enhancements to improve the performance of RAG systems. However, there is limited effort in providing a comprehensive and systematic discussion of the RAG framework. In this work, we conceptualize the RAG framework from three key perspectives: knowledge base, task classification, and system development. We assert that the knowledge base serves as the fundamental cornerstone of RAG, underpinning all retrieval and generation processes. Furthermore, we recognize that RAG tasks can vary significantly in complexity and difficulty, depending on the required generation capabilities and the availability of supporting corpora. By categorizing tasks according to their difficulty levels, we classify RAG systems into distinct levels based on their problem-solving capabilities across the different types of questions.

\subsection{Knowledge base} 

In industrial applications, specialized knowledge primarily originates from years of accumulated data within specific fields such as manufacturing, energy, and logistics. For example, in the pharmaceutical industry, data sources include extensive research and development documentation, as well as drug application files amassed over many years. These sources are not only diverse in file formats, but also encompass a significant amount of multi-modal contents such as tables, charts, and figures, which are also crucial for problem-solving. Furthermore, there are often functional connections between files within a specialized domain, such as hyperlinks, references, and relational database links, which explicitly or implicitly reflect the logical organization of knowledge within the professional field. Currently, existing datasets provide pre-segmented corpora and do not account for the complexities encountered in real-world applications, such as the integration of multi-format data and the maintenance of referential relationships between documents. Therefore, the construction of a comprehensive knowledge base is foundational for Retrieval-Augmented Generation (RAG) in the industrial field. 
As the architecture and quality of the knowledge base directly influence the retrieval methods and their performance, we propose structuring the knowledge base as a multi-layer heterogeneous graph, denoted as $G$, with corresponding nodes and edges represented by $(V,E)$. The graph nodes can include documents, sections, chunks, figures, tables, and customized nodes from distilled knowledge. The edges signify the relationships among these nodes, encapsulating the interconnections and dependencies within the graph. This multi-layer heterogeneous graph encompasses three distinct layers: the information resource layer $G_i$, the corpus layer $G_c$ and the distilled knowledge layer $G_{dk}$. Each layer corresponds to different stages of information processing, representing varying levels of granularity and abstraction in knowledge.

\subsection{Task classification} \label{subsec:task_classification}

Contemporary RAG frameworks frequently overlook the intricate difficulty and logistical demands inherent to diverse tasks, typically employing a one-size-fits-all methodology. However, even with comprehensive knowledge retrieval, current RAG systems are insufficient to handle tasks of varying difficulty with equal effectiveness. Therefore, it is essential to categorize tasks and analyze the typical strategies for overcoming the challenges inherent to each category. 
The difficulty of a task is closely associated with several critical factors.  
\begin{itemize}[left=12pt]
    \item \textit{Relevance and Completeness of Knowledge}: The extent to which the necessary information is present within the knowledge base and how comprehensively it covers the topic.
    \item \textit{Complexity of Knowledge Extraction}: The difficulty in accurately identifying and retrieving all relevant pieces of knowledge, especially when scattered across multiple sources or implicitly embedded in the text.
    \item \textit{Depth of Understanding and Reasoning}: The level of cognitive and inferential processing required to comprehend the retrieved information, establish connections, and perform multi-step reasoning.
    \item \textit{Effectiveness of Knowledge Utilization}: The sophistication involved in applying the extracted knowledge to formulate responses, including synthesizing, organizing, and generating insights or predictions.

\end{itemize}  

\begin{figure}[t]
	\begin{center}
		\includegraphics[width=0.93\linewidth]{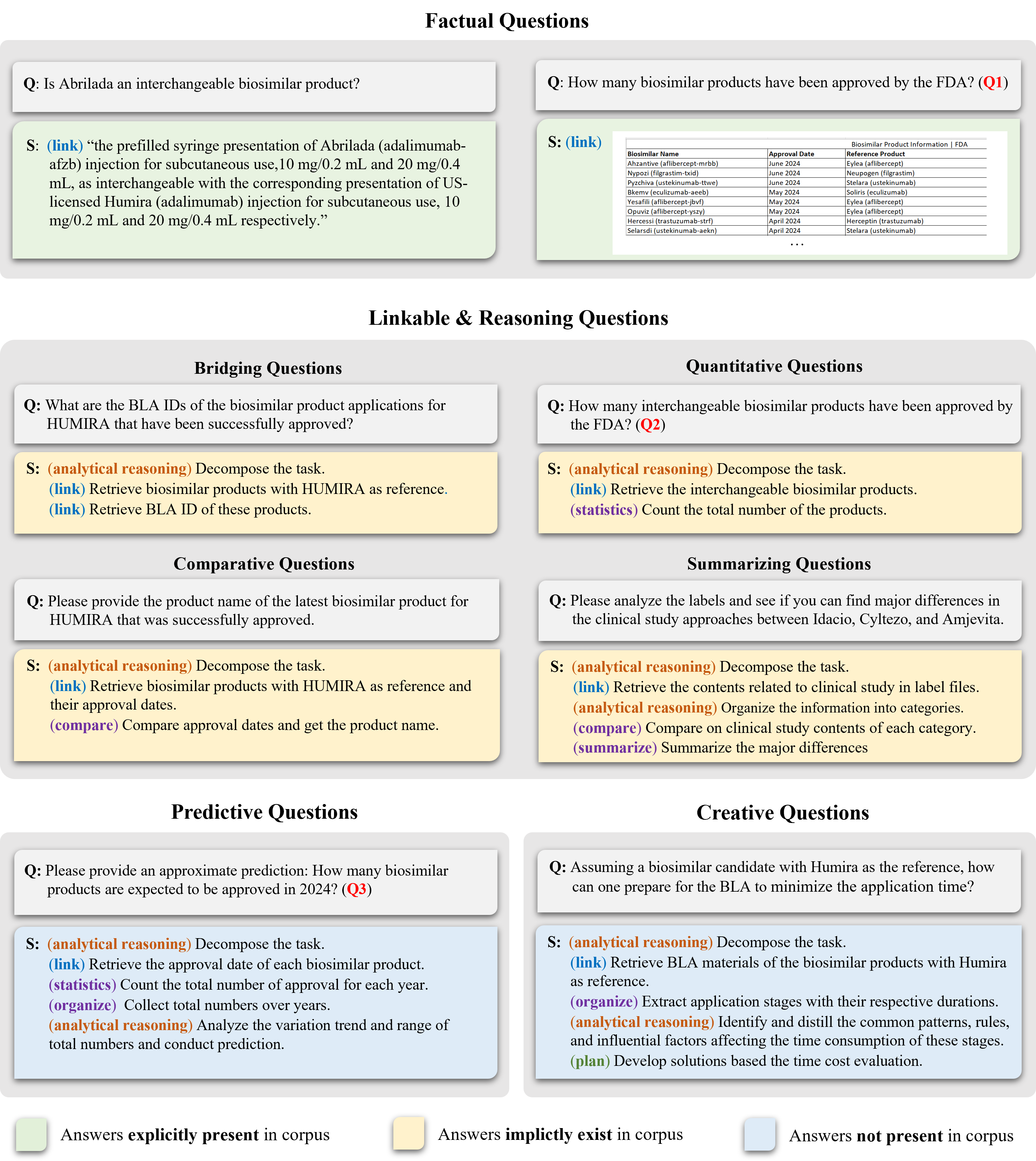}
	\end{center}
	\vspace{-2mm}
	\caption{Illustrative examples of distinct question types} 
	\label{fig-qeustion-type}
	\vspace{-2mm}
\end{figure}

In categorizing real-world RAG tasks within industries, we focus on the processes of knowledge extraction, understanding, organization, and utilization to provide structured and insightful responses. Taking the aforementioned factors into account, we identify four distinct classes of questions that address a broad spectrum of demands. The first type, \textit{Factual Questions}, involves extracting specific, explicit information directly from the corpus, relying on retrieval mechanisms to identify the relevant facts. \textit{Linkable-Reasoning Questions} demand a deeper level of knowledge integration, often requiring multi-step reasoning and linking across multiple sources. 
\textit{Predictive Questions} extend beyond the available data, requiring inductive reasoning and structuring of retrieved facts into analyzable forms, such as time series, for future-oriented predictions. Finally, \textit{Creative Questions} engage domain-specific logic and creative problem-solving, encouraging the generation of innovative solutions by synthesizing knowledge and identifying patterns or influencing factors. This categorization, driven by varying levels of reasoning and knowledge management, ensures a comprehensive approach to addressing industry-specific queries.

The criteria defining each category are elaborated in the following sections, with representative examples for each provided in Figure~\ref{fig-qeustion-type}. For each question type, we also present the associated support data and the expected reasoning processes to illustrate the differences between these categories. These inquiries are formulated by experts in pharmaceutical applications, based on the data released by the FDA.\footnote{\url{https://www.fda.gov/}}


\paragraph{Factual Questions}
These questions seek specific, concrete pieces of information explicitly presented in the original corpus. The referenced text can be processed within the context of a conversation in LLMs. As shown in Figure~\ref{fig-qeustion-type}, this class of questions can be effectively answered if the relevant fact is successfully retrieved.

\paragraph{Linkable-Reasoning Questions} 
Answering these questions necessitates gathering pertinent information from diverse sources and/or executing multi-step reasoning. The answers may be implicitly distributed across multiple texts. Due to variations in the linking and reasoning processes, we further divide this category into four subcategories: bridging questions, comparative questions, quantitative questions, and summarizing questions. Examples of each subcategory are illustrated in Figure~\ref{fig-qeustion-type}. 
Specifically, bridging questions involve sequentially bridging multiple entities to derive the answer. Quantitative questions require statistical analysis based on the retrieved data. Comparative questions focus on comparing specified attributes of two entities. Summarizing questions require condensing or synthesizing information from multiple sources or large volumes of text into a concise, coherent summary, and they often involve integrating key points, identifying main themes, or drawing conclusions based on the aggregated content. Summarizing questions may combine elements of other question types, such as bridging, comparative, or quantitative questions, as they frequently require the extraction and integration of diverse pieces of information to generate a comprehensive and meaningful summary. Given these questions require multi-step retrieval and reasoning, it is crucial to establish a reasonable operation route for answer-seeking in interaction with the knowledge base.

\paragraph{Predictive Questions} 
For this type of questions, the answers are not directly available in the original text and may not be purely factual, necessitating inductive reasoning and prediction based on existing facts. To harness the predictive capabilities of LLMs or other external prediction tools, it is essential to gather and organize relevant knowledge to generate structured data for further analysis. For instance, as illustrated in Figure~\ref{fig-qeustion-type}, all biosimilar products with the approval dates are retrieved, and the total number of approvals for each year is calculated and organized to year-indexed time series data for prediction purposes. Furthermore, it is important to note that the correct answer to predictive questions may not be unique, reflecting the inherent uncertainty and variability in predictive tasks.

\paragraph{Creative Questions} 
One significant demand of RAG is to mine valuable domain-specific logic from professional knowledge bases and introduce novel perspectives that can innovate and advance existing solutions. Addressing creative questions necessitates creative thinking based on the availability of factual information and an understanding of the underlying principles and rules. As illustrated in the example, it is essential to organize the extracted information to highlight key stages and their duration, and then identify common patterns and influential factors. Subsequently, solutions are developed with the objective of evaluating potential outcomes and stimulating fresh ideas. The goal of these responses is to inspire experts to generate innovative ideas, rather than to provide ready-to-implement solutions.
 

It is crucial to recognize that the classification of a question may shift with changes in the knowledge base. Questions Q1, Q2, and Q3 in Figure\ref{fig-qeustion-type}, although seemingly similar, are categorized differently depending on the availability of information and the logical steps required to derive an answer. For instance, Q1 is classified as a factual question because it can be directly answered using a table that concisely lists all biosimilar products along with their respective approval dates, providing sufficient explicit information. In contrast, Q2, which inquires about the total count of interchangeable biosimilar products, cannot be resolved by directly referencing a single explicit source. To answer Q2, one must identify all the products meeting the specified criteria and subsequently calculate the total, necessitating an additional step of statistical aggregation. Therefore, Q2 is categorized as a linkable-reasoning question due to the need for an intermediate processing. Finally, Q3 poses a challenge because the answer does not explicitly exist within the knowledge base. Addressing Q3 requires gathering relevant data, organizing it to infer hidden patterns, and making predictions based on these inferred rules. As a result, Q3 is categorized as a predictive question, indicating the requirement to extrapolate beyond the existing data to forecast potential outcomes or trends.

\begin{table}[t] 
  \caption{Level definition based on RAG system's capability}
  \label{Qtype-table}
  \centering
  \begin{tabular}{c m{9cm}}  

  \toprule
    \scriptsize \textbf{Level}    & \multicolumn{1}{c}{\scriptsize \textbf{System capability description}} \\

    \midrule  
    \scriptsize \textbf{L1}     & \scriptsize The L1 system is designed to provide accurate and reliable answers to \textbf{factual questions}, ensuring a solid foundation for basic information retrieval. 
    \\ 
    \hline 
    
    \scriptsize \textbf{L2}     & \scriptsize The L2 system extends its functionality to include accurate and reliable responses to both \textbf{factual questions} and \textbf{linkable-reasoning questions}, enabling more complex multi-step retrieval and reasoning tasks. 
    \\ 
    \hline

    \scriptsize \textbf{L3}     & \scriptsize The L3 system further enhances its capabilities by incorporating the ability to deliver reasonable predictions for \textbf{predictive questions}, while maintaining accuracy and reliability in answering both \textbf{factual questions} and \textbf{linkable-reasoning questions}.
    \\ 
    \hline

    \scriptsize \textbf{L4}     & \scriptsize The L4 system is capable of proposing well-reasoned plans or solutions to \textbf{creative questions}. In addition, it retains the ability to provide reasonable predictions for \textbf{predictive questions}, alongside accurate and reliable answers to \textbf{factual questions} and \textbf{linkable-reasoning questions}.    
    \\ 
    \bottomrule
  \end{tabular}
\end{table}

\subsection{RAG system level}


 In industrial RAG systems, inquiries encompass a broad spectrum of difficulties and are approached from diverse perspectives. Although RAG systems can leverage the general question-answering(QA) abilities of LLMs, their limited comprehension of expert-level knowledge often leads to inconsistent response quality across questions of varying complexities. 
In response to this status quo, we propose categorizing RAG systems into four distinct levels based on their problem-solving capabilities across the four classes of questions outlined in the previous subsection. This stratified approach facilitates the phased development of RAG systems, allowing capabilities to be incrementally enhanced through iterative module refinement and algorithmic optimization. Our framework is strategically designed to provide a standardized, objective methodology for developing RAG systems that effectively meet the specialized needs of various industry scenarios. 
The definition of RAG systems in different level is presented in Table~\ref{Qtype-table}. It highlights the systems' capabilities to handle increasingly complex queries, demonstrating the evolution from simple information retrieval to advanced predictive and creative problem-solving. Each level represents a step towards more sophisticated interactions with knowledge bases, requiring the RAG systems to demonstrate higher levels of understanding, reasoning, and innovation. 

\begin{figure}[t]
	\begin{center}
		\includegraphics[width=1.0\linewidth]{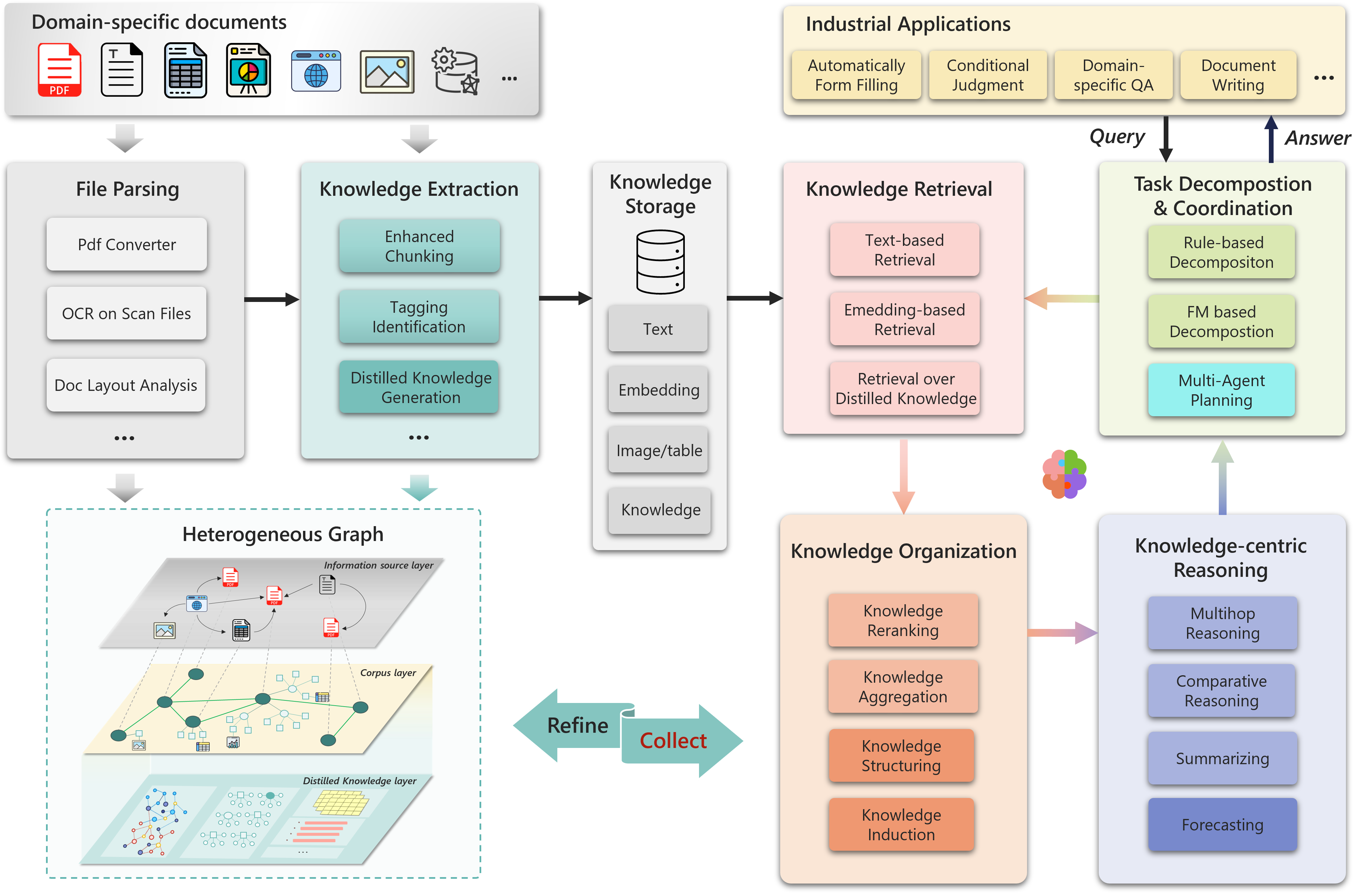}
	\end{center}
	\caption{Overview of the PIKE-RAG framework, comprising several key components: file parsing, knowledge extraction, knowledge storage, knowledge retrieval, knowledge organization, task decomposition and coordination, and knowledge-centric reasoning. Each component can be tailored to meet the evolving demands of system capability.} 
	\label{fig-overview framework}
	\vspace{-5mm}
\end{figure}

 More specially, at the foundational level, RAG systems respond to factual questions with answers that are directly extractable from provided texts. 
 Advancing to the second level, RAG systems are equipped to handle complex questions involving linkage and reasoning. These queries necessitate the synthesis of information from disparate sources or multi-step reasoning processes. The RAG could address a variety of composite questions, includes bridging questions that necessitate a sequence of logical reasoning, comparative questions demanding parallel analysis, and summarizing questions that involve condensing information into comprehensive responses. 
 At the third level, the systems are intricately designed to tackle predictive questions where answers are not immediately discernible from the original text. 
 Finally, RAG systems at the forth level demonstrate the capacity for creative problem-solving, utilizing  a solid factual base to foster novel concepts or strategies. While these systems may not offer ready-to-implement solutions, they play a crucial role in stimulating expert creativity to advance fields such as analytics or treatment design.

\begin{table}[t] 
  \caption{Proposed frameworks for different system levels. To address the challenges facing at each level, we propose customized frameworks based on the framework illustrated in Figure~\ref{fig-overview framework}. The following abbreviations are used: "PA" for file parsing, "KE" for knowledge extraction, "RT" for knowledge retrieval, "KO" for knowledge organization, and "KR" for knowledge-centric reasoning.}
 
  \label{Syslevel-table-figure}
  \centering
  \vspace{5mm}
  \begin{tabular}{|c|m{8cm}|m{3cm}|}  
  \hline  
    \scriptsize \textbf{Level}    & \multicolumn{1}{c|}{\scriptsize  \textbf{Challenges}}     &  \multicolumn{1}{c|}{\scriptsize  \textbf{Proposed Framework}} \\
    \hline

    \small \scriptsize 
    \textbf{L0}
    & \scriptsize 
   \begin{itemize}[left=0pt] 
    \item Challenges arise in \textbf{knowledge extraction} due to the diverse formats of source documents, requiring sophisticated file parsing techniques.  
    \item The construction of a high-quality knowledge base from raw, heterogeneous data introduces significant complexity in \textbf{knowledge organization and integration}. 
    \end{itemize}   
    & 
    \makecell[l]{\includegraphics[width=0.2\textwidth]{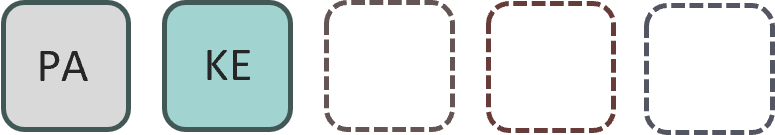}} 
    \\ [-1ex] \hline
    
    \small \scriptsize 
    \textbf{L1}
    & \scriptsize 
   \begin{itemize}[left=0pt] 
    \item The \textbf{understanding and extraction of knowledge} are hindered by improper chunking, which disrupts semantic coherence, complicating accurate retrieval.
    \item \textbf{Knowledge retrieval} is impacted by the limitations of embedding models in aligning professional terminologies and aliases, reducing the system's precision.
    \end{itemize}   
    & 
    \makecell[l]{\includegraphics[width=0.2\textwidth]{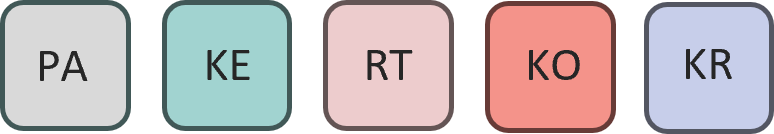}} 
    \\ 
    \hline
    
    \small \scriptsize 
    \textbf{L2}
    & \scriptsize  
    \begin{itemize}[left=0pt]
    \item Effective \textbf{knowledge extraction and utilization} are critical, as chunked text often contains both relevant and irrelevant information. Ensuring the retrieval of high-quality data is essential for accurate generation. 
    \item The \textbf{understanding and decomposition} of tasks and \textbf{rationale} behind them often overlook the availability of supporting data, relying heavily on LLM capabilities.
    \end{itemize}    
    & \vspace{-2mm}
    \makecell[l]{\includegraphics[width=0.2\textwidth]{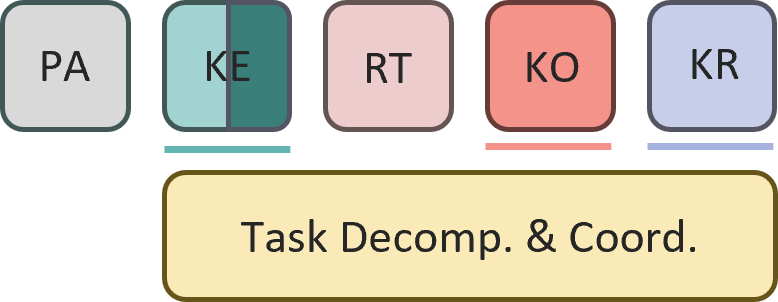}} 
    \\ [-1ex]\hline

    \small \scriptsize 
    \textbf{L3}
    & \scriptsize 
    \begin{itemize}[left=0pt]
    \item The challenges at this level center on \textbf{knowledge collection and organization}, which are vital for supporting predictive reasoning.
    \item LLMs have limitations in applying \textbf{specialized reasoning logic}, restricting their effectiveness in predictive tasks. 
    \end{itemize}        
    & \vspace{-1mm}
    \makecell[l]{\includegraphics[width=0.2\textwidth]{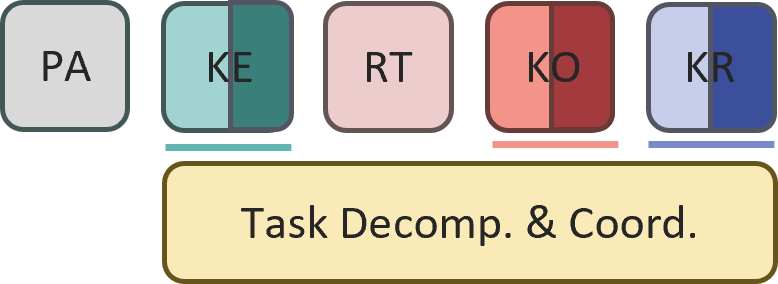}} 
    \\ [-1ex]\hline

    \small \scriptsize
    \textbf{L4}
    & \scriptsize 
    \begin{itemize}[left=0pt] 
    \item The difficulty lies in \textbf{extracting coherent logical rationales} from complex knowledge bases, where interdependencies between multiple factors can result in non-unique solutions.  
    \item The open-ended nature of creative questions complicates the \textbf{evaluation of the reasoning and knowledge synthesis} process, making it difficult to quantitatively assess answer quality. 
    \end{itemize}       
    & \vspace{-2mm}
    \makecell[l]{\includegraphics[width=0.2\textwidth]{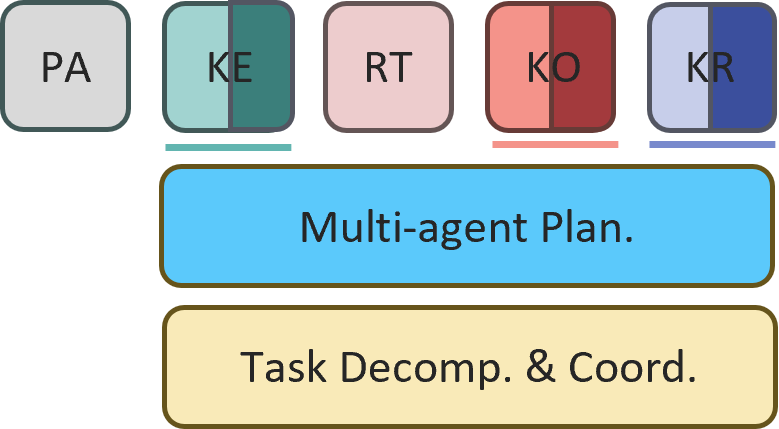}} 
    \\ \hline
  \end{tabular}
\end{table}

\section{Methodology}

\subsection{Framework}
Based on the formulation of RAG systems in terms of knowledge base, task classification, and system-level division, we propose a versatile and expandable RAG framework. Within this framework, the progression in levels of RAG systems can be achieved by adjusting submodules within the main modules. The overview of our framework is depicted in Figure~\ref{fig-overview framework}. The framework primarily consists of several fundamental modules, including file parsing, knowledge extraction, knowledge storage, knowledge retrieval, knowledge organization, knowledge-centric reasoning, and task decomposition and coordination. 
In this framework, domain-specific documents of diverse formats are processed by file parsing module to convert the file to machine-readable formats, and file units are generated to build up graph in information source layer. The knowledge extraction module chunks the text and generates corpus and knowledge units to construct graph in corpus layer and distilled knowledge layer. The heterogeneous graph established is utilized as the knowledge base for retrieval. Extracted knowledge is stored in multiple structured formats, and the knowledge retrieval module employs hybrid retrieval strategy to access relevant information. Note that the knowledge base not only serves as the source of knowledge gathering but also benefits from a feedback loop, where the organized and verified knowledge is regarded as feedback to refine and improve the knowledge base.

As highlighted in the task classification examples, questions of different classes require distinct rationale routing for answer-seeking, influenced by multiple factors such as the availability of relevant information, the complexity of knowledge extraction, and the sophistication of reasoning. It is challenging to address these questions in a single retrieval and generation pass. To tackle this, we propose an iterative retrieval-generation mechanism supervised by task decomposition and coordination. This iterative mechanism enables the gradual collection of relevant information and progressive reasoning over incremental context, ensuring a more accurate and comprehensive response. More specially, the questions in industrial applications are fed into task decomposition module to produce preliminary decomposition scheme. This scheme outlines the retrieval steps, reasoning steps, and other necessary operations. Following these instructions, the knowledge retrieval module retrieves relevant information, which is then passed to the knowledge organization module for processing and organization. The organized knowledge is used to perform knowledge-centric reasoning, yielding an intermediate answer. With the updated relevant information and intermediate answer, the task decomposition module regenerates an updated scheme for the next iteration. This design boasts excellent adaptability, allowing us to tackle problems of varying difficulties and perspectives by adjusting the modules and iterative mechanisms.


\subsection{Phased system development}

We have categorized RAG systems into four distinct levels based on their problem-solving capabilities across the four classes of questions, as outlined in Table~\ref{Qtype-table}. Recognizing the pivotal role of knowledge base generation in RAG systems, we designate the construction of the knowledge base as the L0 stage of system development. The challenges faced by RAG systems vary across different levels. We analyze these challenges for each level and propose corresponding frameworks in Table~\ref{Syslevel-table-figure}. This stratified approach facilitates the phased development of RAG systems, enabling incremental enhancement of capabilities through iterative module refinement and algorithmic optimization. 

We observe that from L0 to L4, higher-level systems can inherit modules from lower levels and add new modules to enhance system capabilities. For instance, compared to an L1 system, an L2 system not only introduces a task decomposition and coordination module to leverage iterative retrieval-generation routing but also incorporates more advanced knowledge extraction modules, such as distilled knowledge generation, indicated in dark green in Figure~\ref{fig-overview framework}. In the L3 system, the growing emphasis on predictive questioning necessitates enhanced requirements for knowledge organization and reasoning. Consequently, the knowledge organization module introduces additional submodules for knowledge structuring and knowledge induction, indicated in dark orange. Similarly, the knowledge-centric reasoning module has been expanded to include a forecasting submodule, highlighted in dark purple. In the L4 system, extracting complex rationale from an established knowledge base is highly challenging. To address this, we introduce multi-agent planning module to activate reasoning from diverse perspectives.

\begin{figure}[t]
	\begin{center}
		\includegraphics[width=0.7\linewidth]{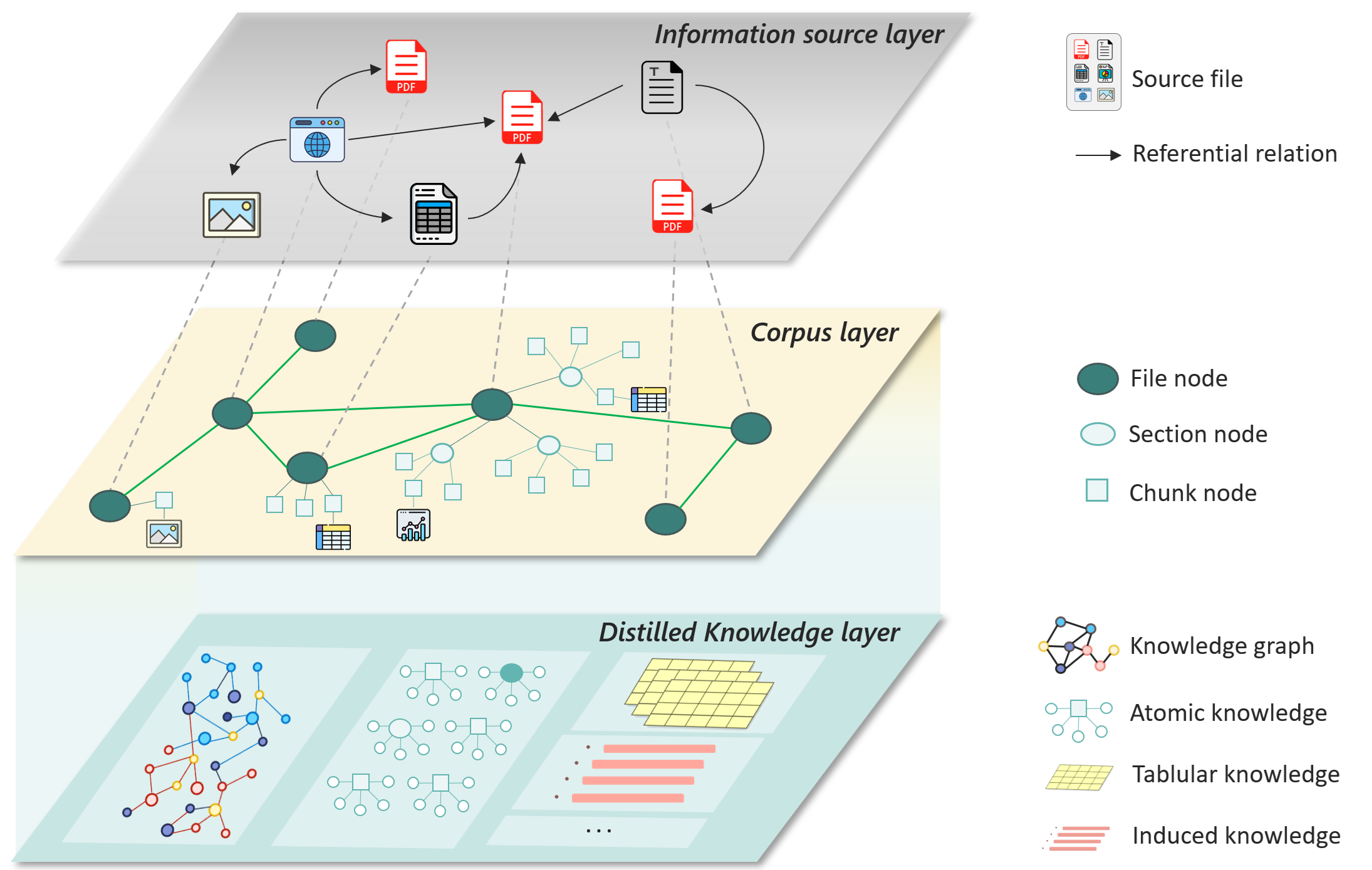}
	\end{center}
	\caption{Multi-layer heterogeneous graph as the knowledge base. The graph comprises three distinct layers: information resource layer, corpus layer and distilled knowledge layer.} 
	\label{fig-graph}
	\vspace{-2mm}
\end{figure}

\section{Detailed Implementation}

In this section, we delve into the implementation specifics of each module within our proposed versatile and expandable RAG framework. By elucidating the details at each level, we aim to provide a comprehensive understanding of how the framework operates and how its modularity and expandability are achieved. The subsections that follow will cover the file parsing, knowledge extraction, knowledge storage, knowledge-centric reasoning, and task decomposition and coordination modules, providing insights into their individual functionalities and interactions.

\subsection{Level-0: Knowledge Base Construction}
The foundational stage of the proposed RAG systems is designated as the L0 system, focuses on the construction of a robust and comprehensive knowledge base. This stage is critical for enabling effective knowledge retrieval in subsequent levels. The primary objective of the L0 system is to process and structure domain-specific documents, transforming them into a machine-readable format and organizing the extracted knowledge into a heterogeneous graph. This graph serves as the backbone for all higher-level reasoning and retrieval tasks.
The L0 system encompasses several key modules: file parsing, knowledge extraction, and knowledge storage. Each of these modules plays a crucial role in ensuring that the knowledge base is both extensive and accurately reflects the underlying information contained within the source documents.


\subsubsection{File parsing}
The ability to effectively parse and read various types of files is a critical component in the development of RAG systems that rely on diverse data sources. Frameworks such as LangChain\footnote{https://www.langchain.com} provide a comprehensive suite of tools for natural language processing (NLP), including modules for parsing and extracting information from unstructured text documents. Its file reader capabilities are designed to handle a wide range of file formats, ensuring that data from heterogeneous sources can be seamlessly integrated into the system. Additionally, several deep learning-based tools \cite{paddlepaddle,tesseract} and commercial cloud APIs~\cite{documentintelligence,Textract} have been developed to conduct robust Optical Character Recognition (OCR) and accurate table extraction, enabling the conversion of scanned documents and images into structured, machine-readable text. Given that domain-specific files often encompass sophisticated tables, charts, and figures, text-based conversion may lead to information loss and disrupt the inherent logical structure. Therefore, we propose conducting layout analysis for these files and preserving multi-modal elements such as charts and figures. The layout information can aid the chunking operation, maintaining the completeness of chunked text, while figures and charts can be described by Vision-Language Models (VLMs) to assist in knowledge retrieval. This approach ensures that the integrity and richness of the original documents are retained, enhancing the efficacy of RAG systems.


\begin{figure}[t]
	\begin{center}
		\includegraphics[width=0.8\linewidth]{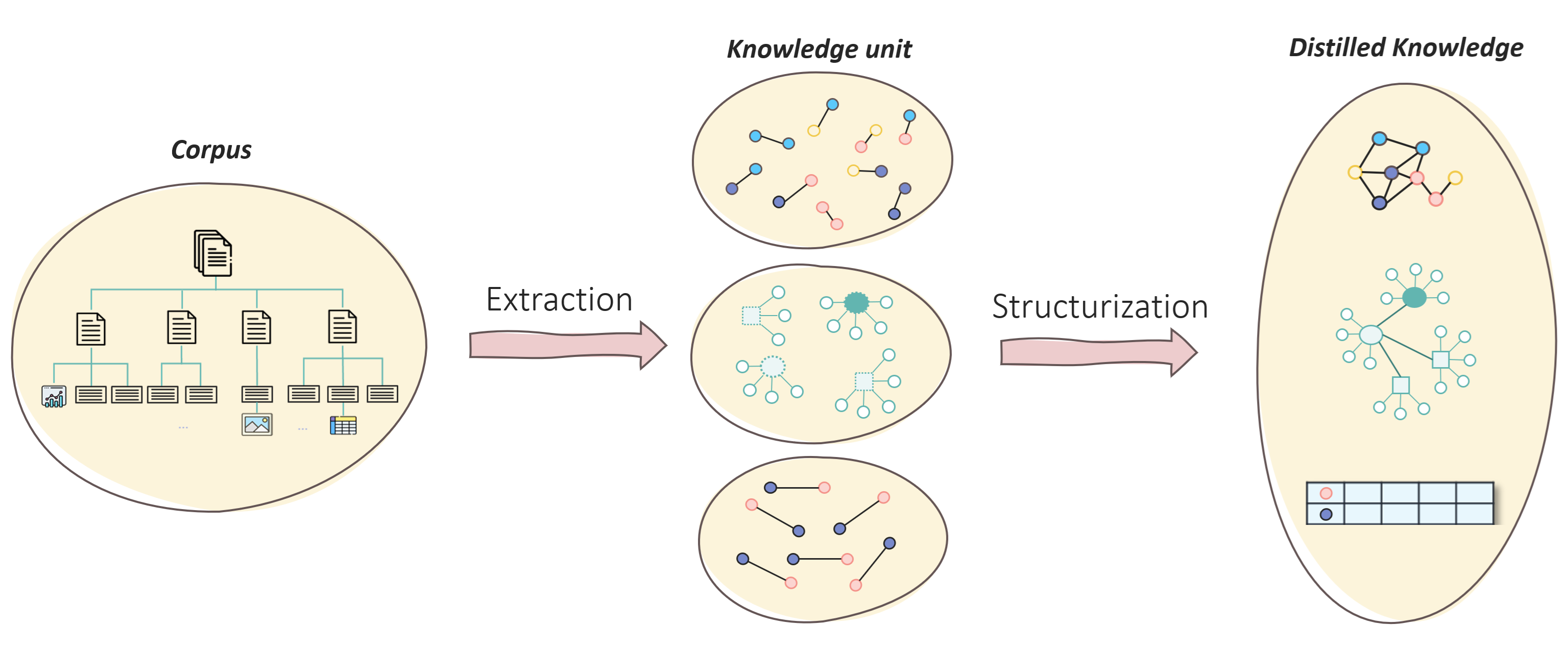}
	\end{center}
	\caption{The process of distilling knowledge from corpus text. The corpus text are processed to extract knowledge units following customized extraction patterns. These knowledge units are then organized to structured knowledge in the distilled knowledge layer, which may take the form of knowledge graphs, atomic knowledge, tabular knowledge, and other induced knowledge.} 
	\label{fig-distill-knowledge}
\end{figure}

\subsubsection{Knowledge Organization} \label{sec:heterogeneours-graph}

The proposed knowledge base is structured as a multi-layer heterogeneous graph, representing different levels of information granularity and abstraction. The graph captures relationships between various components of the data (e.g., documents, sections, chunks, figures, and tables) and organizes them into nodes and edges, reflecting their interconnections and dependencies. As depicted in Figure~\ref{fig-graph}, this multi-layer structure, encompassing the information resource layer, corpus layer, and distilled knowledge layer, enables both semantic understanding and rationale-based retrieval for downstream tasks.

\textit{Information Resource Layer:} This layer captures the diverse information sources, treating them as source nodes with edges that denote referential relationships among them. This structure aids in cross-referencing and contextualizing the knowledge, establishing a foundation for reasoning that depends on multiple sources.

\textit{Corpus Layer:} This layer organizes the parsed information into sections and chunks while preserving the document's original hierarchical structure. Multi-modal content such as tables and figures is summarized by LLMs and integrated as chunk nodes, ensuring that multi-modal knowledge is available for retrieval. This layer enables knowledge extraction with varying levels of granularity, allowing for accurate semantic chunking and retrieval across diverse content types.

\textit{Distilled Knowledge Layer:} The corpus is further distilled into structured forms of knowledge (e.g., knowledge graphs, atomic knowledge, and tabular knowledge). This process, driven by techniques like Named Entity Recognition (NER)~\cite{collobert2011natural} and relationship extraction~\cite{mintz2009distant}, ensures that the distilled knowledge captures key logical relationships and entities, supporting advanced reasoning processes. By organizing this structured knowledge in a distilled layer, we enhance the system's ability to reason and synthesize based on deeper domain-specific knowledge.
The knowledge distillation process is depicted in Figure~\ref{fig-distill-knowledge}. Below are the detailed distillation processes for typical knowledge forms.

\begin{itemize}[left=12pt]  
    \item \textit{Knowledge graph}: Entities and their relationships are extracted from the corpus text using LLMs, generating knowledge units in form of “node-edge-node” structure, where nodes represent entities and edges represent the relationships between them. All knowledge units are then integrated to construct a graph.  
    \item \textit{Atomic knowledge}: The corpus text is partitioned into a set of atomic statements, which are considered as knowledge units. By combining these atomic statements with the relationships between corpus nodes, atomic knowledge is generated.  
    \item \textit{Tabular knowledge}: Entity pairs with specified types and relationships are extracted from corpus text. These entity pairs are treated as knowledge units and can be combined to construct tabular knowledge.  
\end{itemize}  


\begin{figure}[t]
	\begin{center}
		\includegraphics[width=0.8\linewidth]{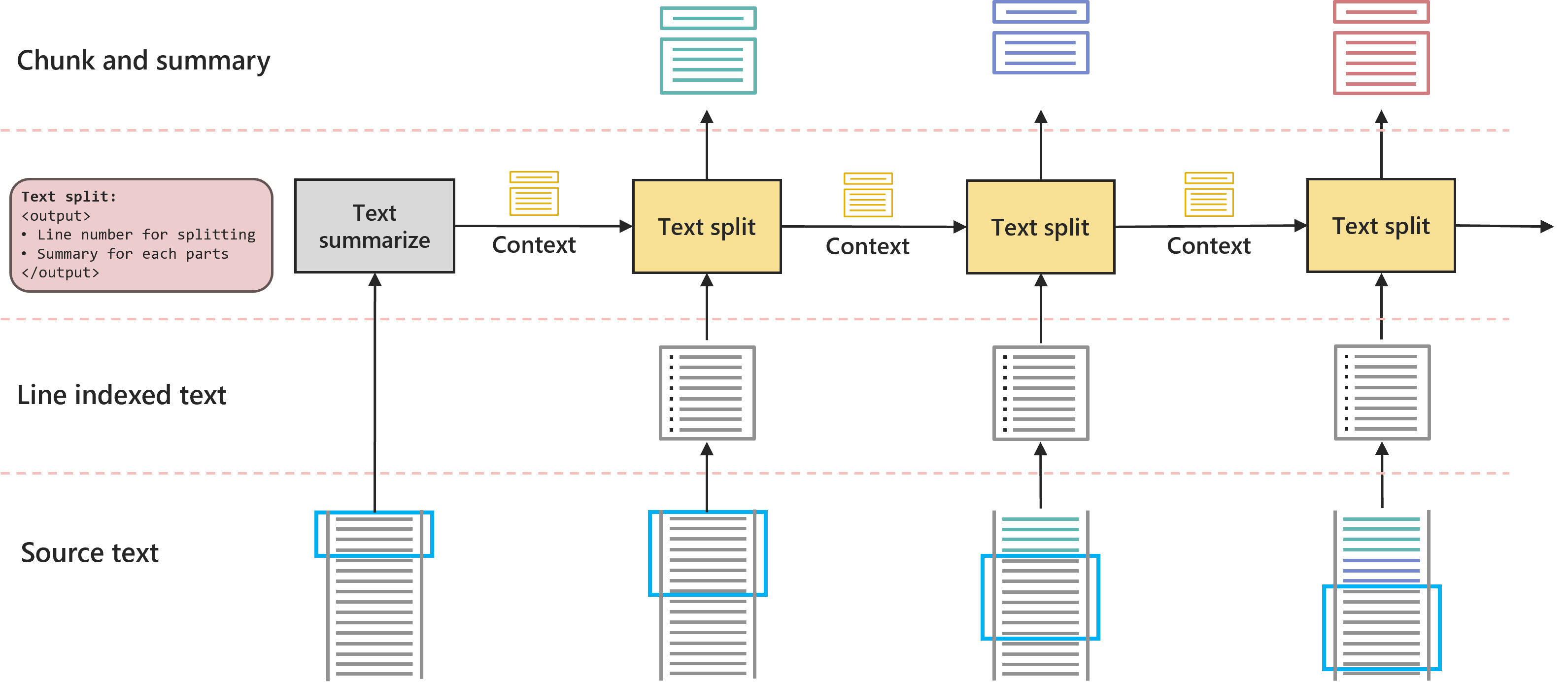}
	\end{center}
	\vspace{-2mm}
	\caption{Illustration of enhanced chunking with recurrent text splitting.} 
	\label{fig-chunking}
\end{figure}

\subsection{Level-1: Factual Question focused RAG System}

Building upon the L0 system, the L1 system introduces knowledge retrieval and knowledge organization to realize its retrieval and generation capabilities. The primary challenges at this level are semantic alignment and chunking. The abundance of professional terminology and aliases can affect the accuracy of chunk retrieval, and unreasonable chunking can disrupt semantic coherence and introduce noise interference. To mitigate these issues, the L1 system incorporates more sophisticated query analysis techniques and basic knowledge extraction modules. The architecture is expanded to include components that facilitate task decomposition, coordination, and initial stages of knowledge organization (KO), ensuring that the system can manage more complex queries effectively.
\begin{figure}[h]
	\begin{center}
		\includegraphics[width=0.7\linewidth]{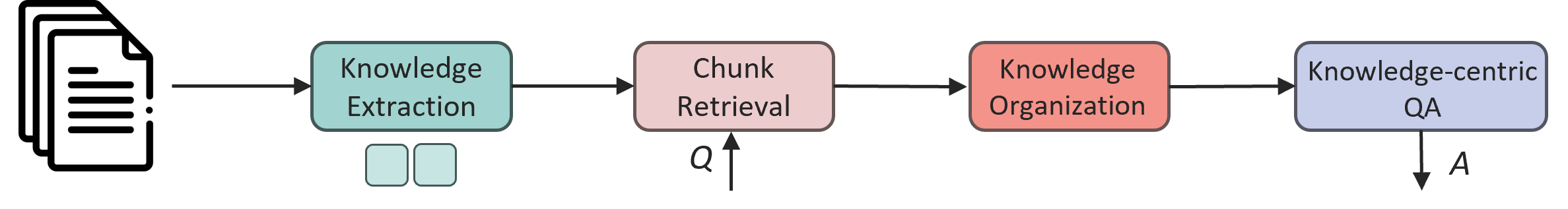}
	\end{center}
	\vspace{-2mm}
	\caption{Overview of L1 RAG framework. The squares $($\includegraphics[height=0.2cm]{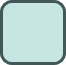}$)$ indicate enhanced chunking and auto-tagging sub-module in knowledge extraction modules.} 
	\label{fig-L1}
	\vspace{-2mm}
\end{figure}

\subsubsection{Enhanced chunking}

Chunking involves breaking down a large corpus of text into smaller, more manageable segments. The primary chunking strategies commonly utilized in RAG systems include fixed-size chunking, semantic chunking, and hybrid chunking. Chunking is essential for improving both the efficiency and accuracy of the retrieval process, which consequently affects the overall performance of RAG models in multiple dimensions. In our system, each chunk serves dual purposes: (i) it becomes a unit of information that is vectorized and stored in a database for retrieval, and (ii) it acts as a source for further knowledge extraction and information summarization. Improper chunking not only fails to ensures that text vectors encapsulate the necessary semantic information, but also hinders knowledge extraction based on complete context. For instance, in the context of laws and regulations, a fixed-size chunking approach are prone to destroying text semantics and omitting key conditions, thereby affecting the quality and accuracy of subsequent knowledge extraction.

We propose a text split algorithm to enhance existing chunking methods by breaking down large text documents into smaller, manageable chunks while preserving context and enabling effective summary generation for each chunk. The chunking process is illustrated in Figure \ref{fig-chunking}. Given a source text, the algorithm iteratively splits the text into chunks. During the first iteration, it generates a forward summary of the initial chunk, providing context for generating summaries of subsequent chunks and maintaining a coherent narrative across splits. Each chunk is summarized using a predefined prompt template that incorporates both the forward summary and the current chunk. This summary is then stored alongside the chunk. The algorithm adjusts the text by removing the processed chunk and updating the forward summary with the summary of the current chunk, preparing for the next iteration. This process continues until the entire text is split and summarized. Additionally, the algorithm can dynamically adjust chunk sizes based on the content and structure of the text.

\begin{figure}[t]
	\begin{center}
		\includegraphics[width=0.85\linewidth]{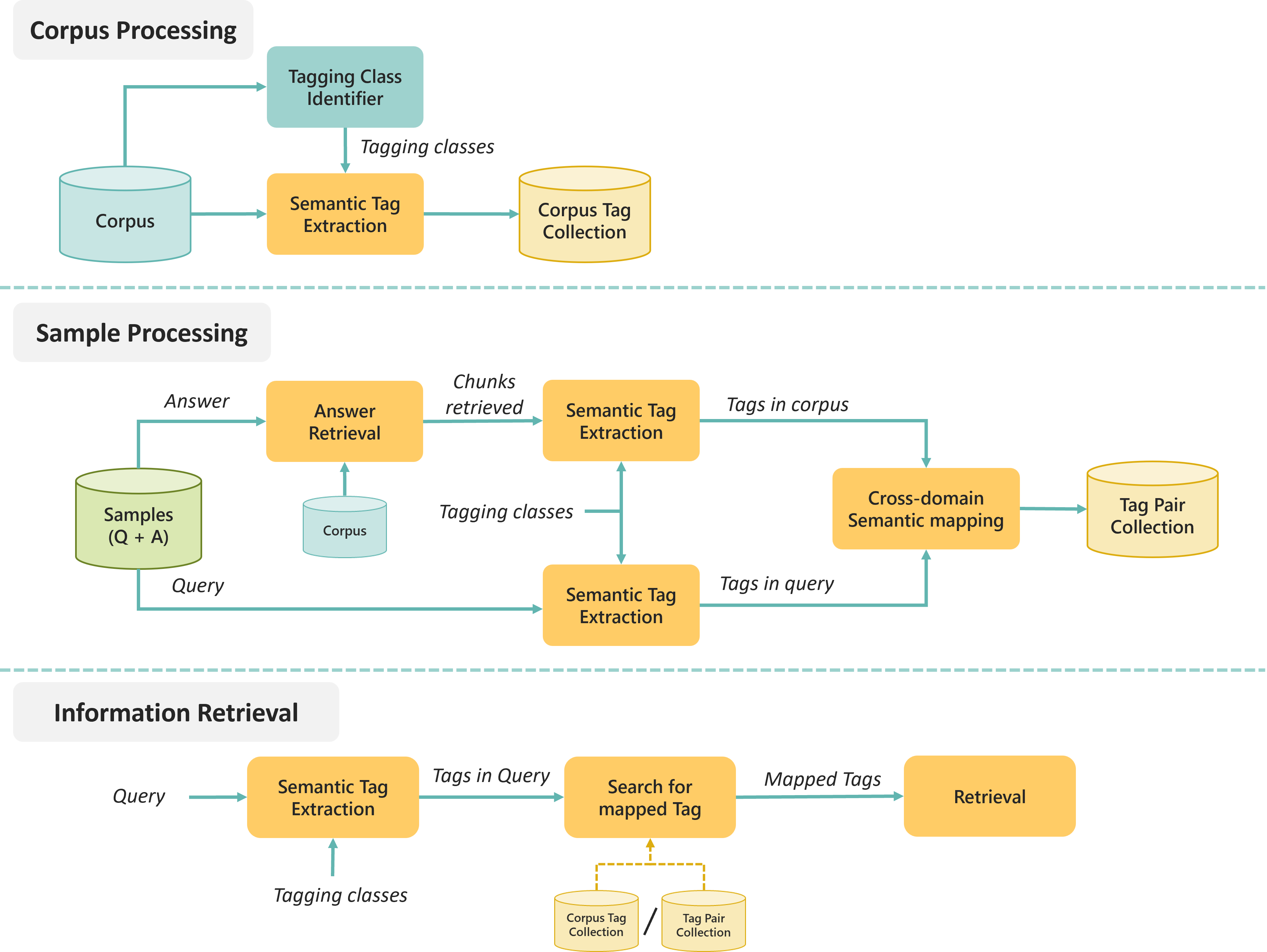}
	\end{center}
	\vspace{-2mm}
	\caption{Illustration of the auto-tagging module.} 
	\label{fig-auto-tagging}
	\vspace{-2mm}
\end{figure}

\subsubsection{Auto-tagging}

In domain-specific RAG scenarios, the corpus is typically characterized by formal, professional, and rigorously expressed content, whereas the questions posed are often articulated in plain, easily understandable colloquial language. For instance, in medical question-answering (medQA) tasks~\cite{jin2020disease}, symptoms of diseases described in the questions are generally phrased in simple, conversational terms. In contrast, the corresponding medical knowledge within the corpus is often expressed using specialized professional terminology. This discrepancy introduces a domain gap that adversely affects the accuracy of chunk retrieval, especially given the limitations of the embedding models employed for this purpose. 

To address the domain gap issue, we propose an auto-tagging module designed to minimize the disparity between the source documents and the queries. This module preprocesses the corpus to extract a comprehensive collection of domain-specific tags or to establish tag mapping rules. Prior to the retrieval process, tags are extracted from the query and then mapped to corpus domain using the preprocessed tag collection or tag pair collection. This tag-based domain adaptation can be employed for query rewriting or keyword retrieval within sequential information retrieval frameworks, thereby enhancing both the recall and precision of the retrieval process. 


Specifically, we leverage the capabilities of the LLMs to identify key factors within the corpus chunks, summarize these factors, and generalize them into category names, which we refer to as "tag classes." We generate semantic tag extraction prompts based on these tag classes to facilitate accurate tag extraction. In scenarios where only the corpus is available, LLMs are employed with meticulously designed prompts to extract semantic tags from the corpus, thereby forming a comprehensive corpus tag collection. When practical QA samples are available, semantic tag extraction is performed on both the queries and the corresponding retrieved answer chunks. Using the tag sets extracted from the chunks and queries, LLMs are utilized to map cross-domain semantic tags and generate a tag pair collection. After establishing both the corpus tag collection and the tag pair collection, tags can be extracted from the query, and the corresponding mapped tags can be identified within the collections. These mapped tags are then used to enhance subsequent information retrieval processes, improving both recall and precision. This workflow leverages the advanced understanding and contextual capabilities of LLMs for domain adaptation.

\subsubsection{Multi-Granularity Retrieval}

The L1 system is designed to enable multi-layer, multi-granularity retrieval across a heterogeneous knowledge graph, which was constructed in the L0 system. Each layer of the graph (e.g., information source layer, corpus layer, distilled knowledge layer) represents knowledge at different levels of abstraction and granularity, allowing the system to explore and retrieve relevant information at various scales. For example, queries can be mapped to entire documents (information source layer) or specific chunks of text (corpus layer), ensuring that knowledge can be retrieved at the appropriate level for a given task.
To support this, similarity scores between queries and graph nodes are computed to measure the alignment between the query and the retrieved knowledge. These scores are then propagated through the layers of the graph, allowing the system to aggregate information from multiple levels. This multi-layer propagation ensures that retrieval can be fine-tuned based on both the broader context (e.g., entire documents) and finer details (e.g., specific chunks or distilled knowledge). 
The final similarity score is generated through a combination of aggregation and propagation, ensuring that knowledge extraction and utilization are optimized for both precision and efficiency in factual question answering. The retrieval process can be iterative, refining the results based on sub-queries generated through task decomposition, further enhancing the system's ability to generate accurate and contextually relevant answers.

\begin{figure}[t]
	\begin{center}
		\includegraphics[width=0.9\linewidth]{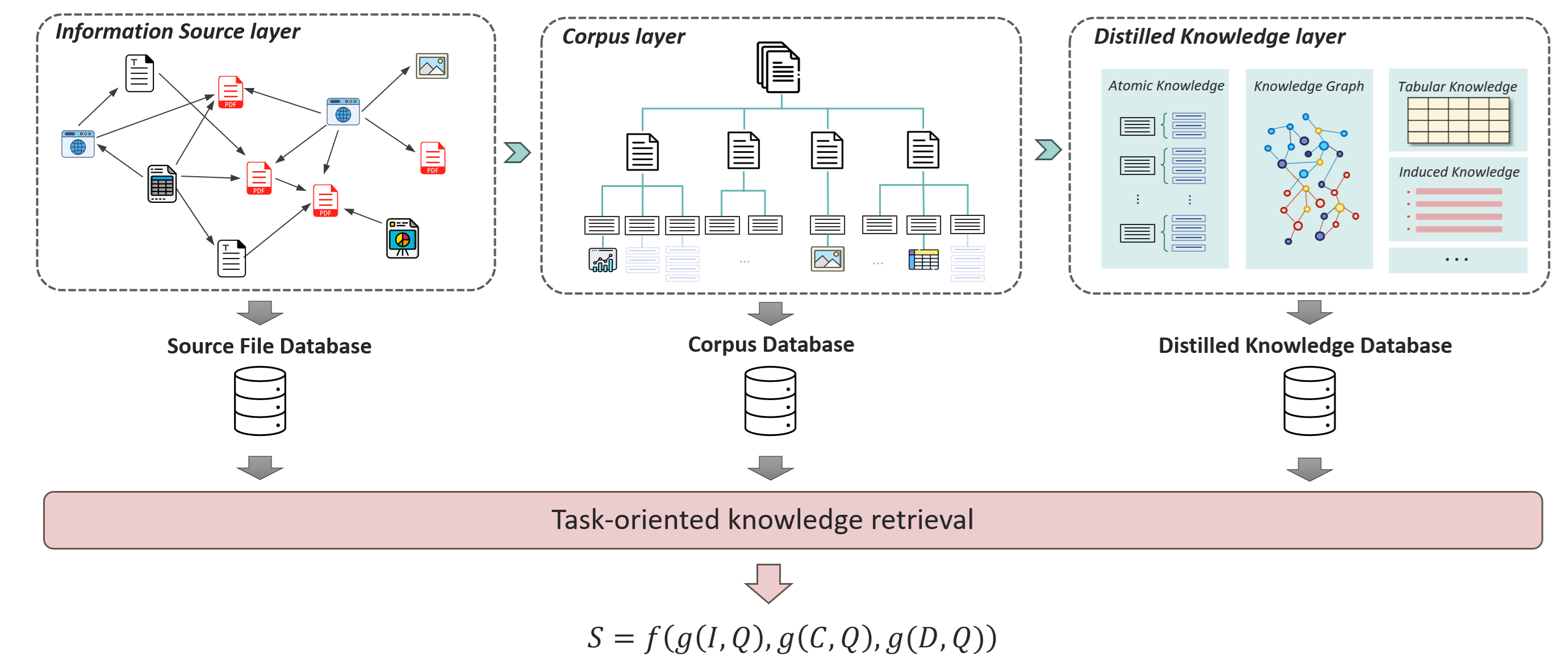}
	\end{center}
	\vspace{-1mm}
	\caption{Overview of multi-layer, multi-granularity retrieval over heterogeneous graph} 
	\label{fig-graph-retrieve}
\end{figure}

The overview of multi-layer, multi-granularity retrieval is depicted in Figure~\ref{fig-graph-retrieve}. For each layer of the graph, both queries $Q$ and graph node are transformed into high-dimensional vector embeddings for similarity evaluation. We denote the similarity evaluation operation as $g(\ast)$. Here, $I$, $C$, and $D$ indicate the node sets in the information source layer, corpus layer, and distilled knowledge layer, respectively. The propagation and aggregation operations are represented by the function $f(\ast)$. The final chunk similarity score $S$ is obtained by aggregating the scores from other layers and nodes. 

\subsection{Level-2: Linkable and Reasoning Question focused RAG System} \label{subsec:l2_atomic_decompo}

\begin{figure}[ht]
	\begin{center}
		\includegraphics[width=0.75\linewidth]{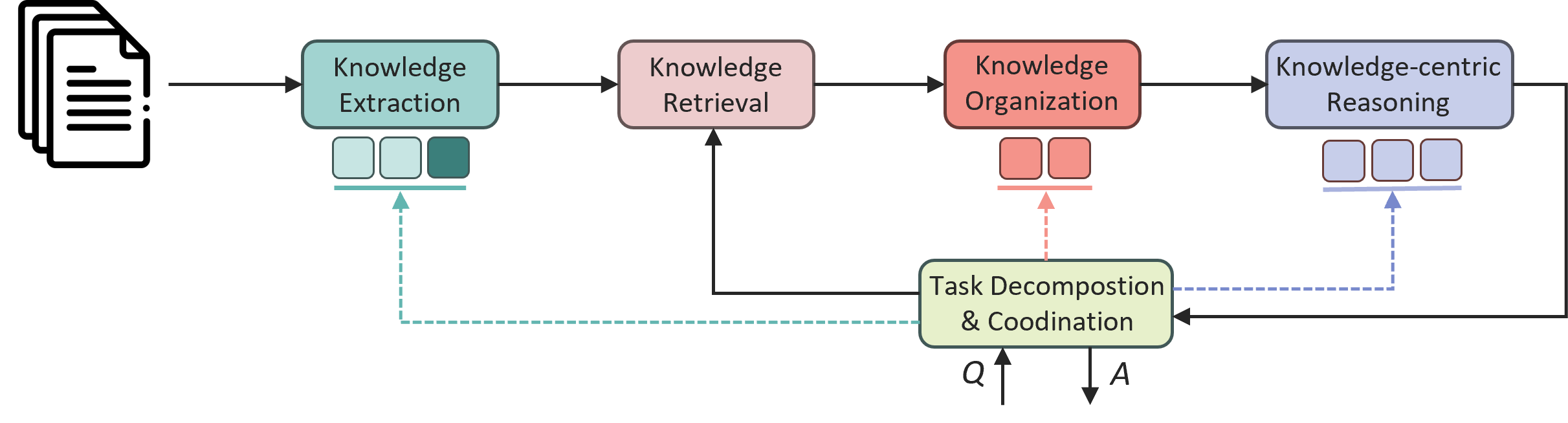}
	\end{center}
	\vspace{-2mm}
	\caption{Overview of L2 RAG framework. The square $($\includegraphics[height=0.2cm]{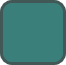}$)$ indicates atomic knowledge generation in knowledge extraction module, while the squares $($\includegraphics[height=0.2cm]{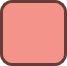}$)$ represent knowledge reranking and aggregation sub-module in knowledge origination module. Moreover, the squares $($\includegraphics[height=0.2cm]{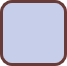}$)$ indicate the multihop reasoning, comparative reasoning, summarizing sub-modules in knowledge-centric reasoning module.} 
	\label{fig-L2}
\end{figure}

The core functionality of the L2 system lies in its ability to efficiently retrieve multiple sources of relevant information and perform complex reasoning based on it. To facilitate this, the L2 system integrates an advanced knowledge extraction module that comprehensively identifies and extracts pertinent information. Furthermore, a task decomposition and coordination module is implemented to break down intricate tasks into smaller, manageable sub-tasks, thereby enhancing the system's efficiency in handling them. The proposed framework of L2 RAG system is illustrated in Figure~\ref{fig-L2}.


Chunked text contains multifaceted information, increasing the complexity of retrieval. Recent studies have focused on extracting triple knowledge units from chunked text and constructing knowledge graphs to facilitate efficient information retrieval~\cite{edge2024localglobalgraphrag,panda2024holmeshyperrelationalknowledgegraphs}. However, the construction of knowledge graphs is costly, and the inherent knowledge may not always be fully explored. To better present the knowledge embedded the documents, we propose atomizing the original documents in Knowledge Extraction phase, a process we refer as \textit{Knowledge Atomizing}.
Besides, industrial tasks often necessitate multiple pieces of knowledge, implicitly requiring the capability to decompose the original question into several sequential or parallel atomic questions. We refer to this operation as \textit{Task Decomposition}. By combining the extracted atomic knowledge with the original chunks, we construct an atomic hierarchical knowledge base. Each time we decompose a task, the hierarchical knowledge base provides insights into the available knowledge, enabling knowledge-aware task decomposition.

\begin{figure}[t]
	\begin{center}
		\includegraphics[width=0.99\linewidth]{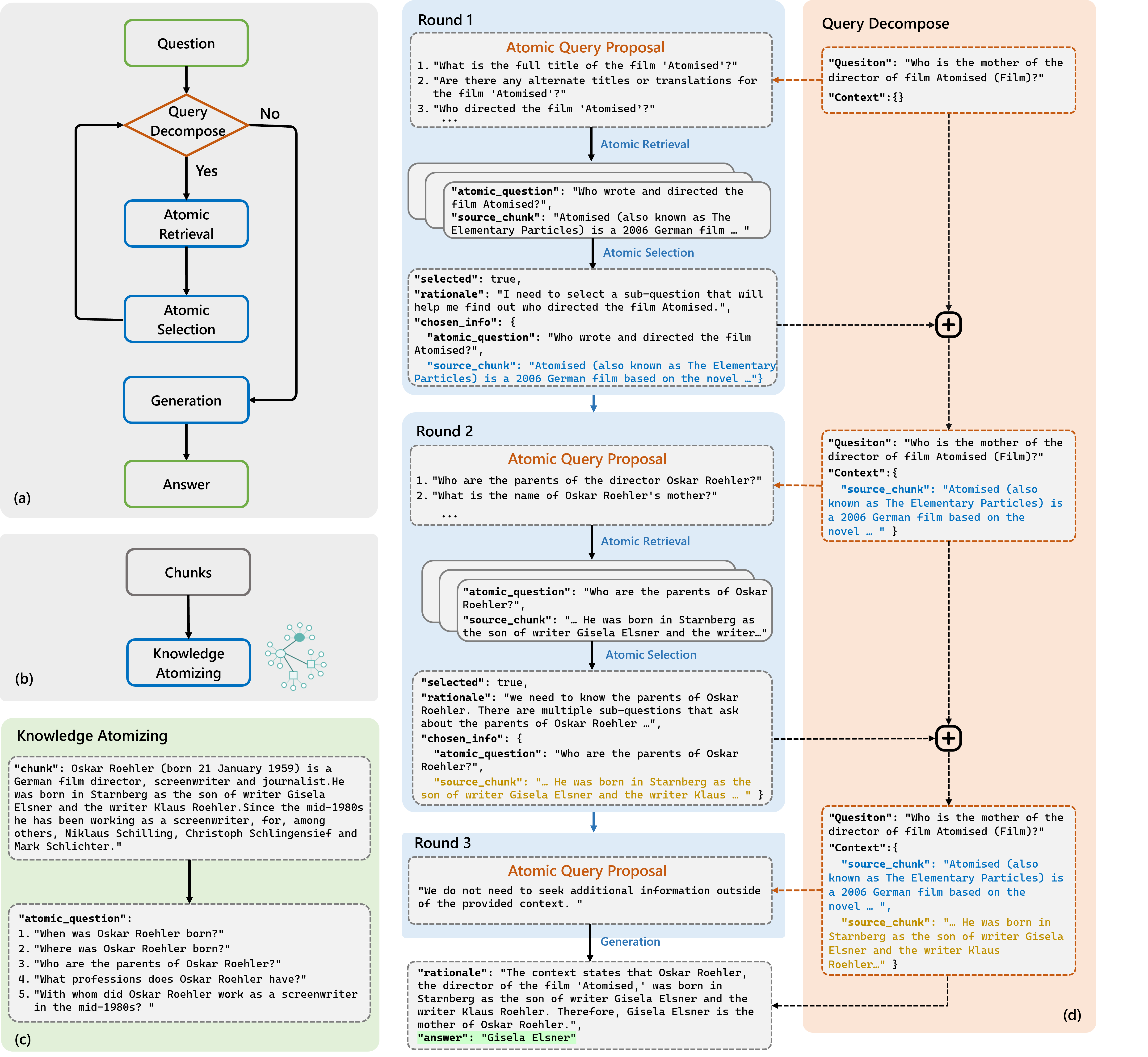}
	\end{center}
	\caption{The illustration of knowledge atomizing and knowledge-aware task decomposition: (a) Workflow of task solving with knowledge-aware task decomposition, (b) Workflow of knowledge atomizing, (c) Example of knowledge atomizing, (d) RAG case with knowledge atomizing and knowledge-aware task decomposition.} 
    \vspace{-4mm}
        \label{fig: atomic-example}
\end{figure}

\subsubsection{Knowledge Atomizing}
We believe that a single document chunk often encompasses multiple pieces of knowledge. Typically, the information necessary to address a specific task represents only a subset of the entire knowledge. Therefore, consolidating these pieces within a single chunk, as traditionally done in information retrieval, may not facilitate the efficient retrieval of the precise information required.
To align the granularity of knowledge with the queries generated during task solving, we propose a method called knowledge atomizing. This approach leverage the context understanding and content generation capabilities of LLMs to automatically tag atomic knowledge pieces within each document chunk. Note that, these chunks could be segments of an original reference document, description chunks generated for tables, images, videos, or summary chunks of entire sections, chapters or even documents.

\begin{figure}[t]
    \begin{center}
        \includegraphics[width=0.6\linewidth]{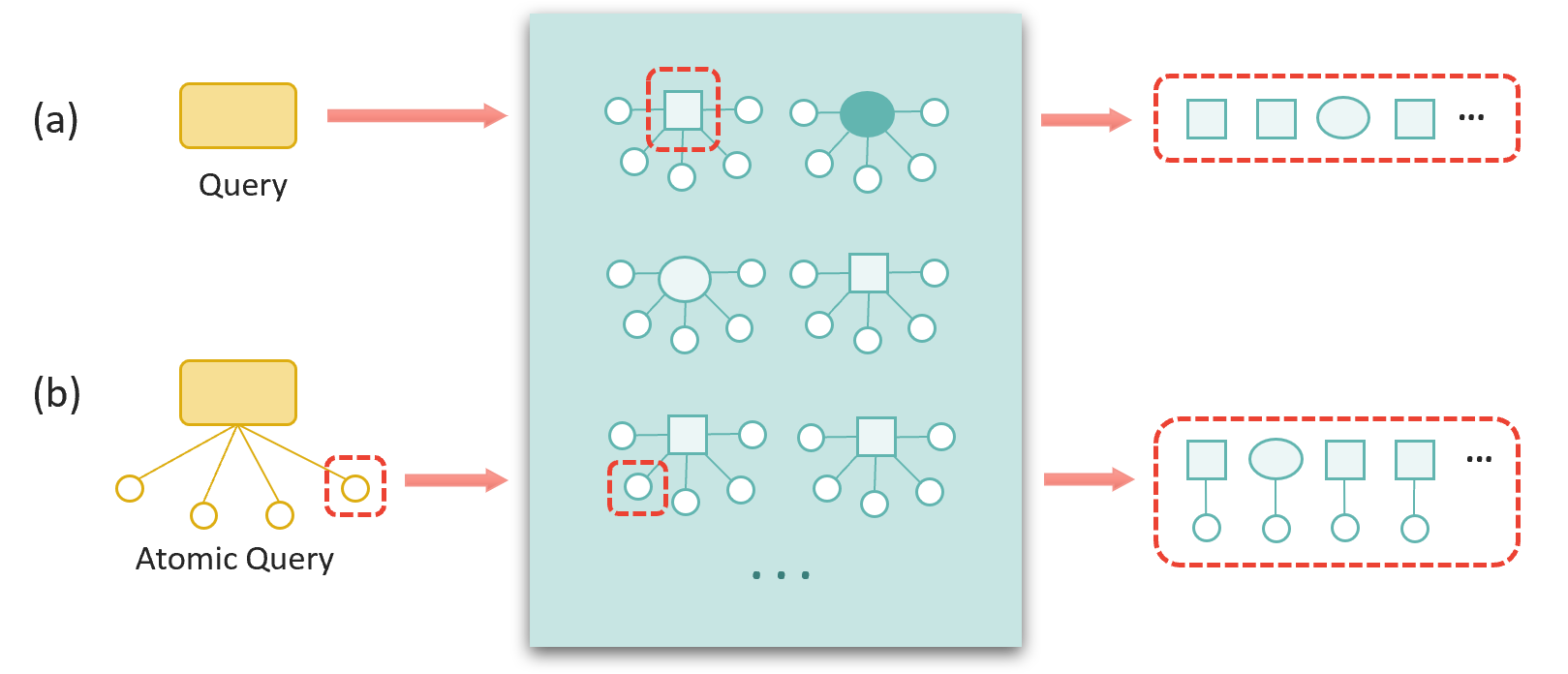}
    \end{center}
    \vspace{-2mm}
    \caption{Retrieval process from an atomic knowledge base. It supports two retrieval paths: (a) using queries to directly retrieve chunks as usual; (b) locating atomic nodes first then achieving the associated chunks.} 
    \label{fig: 2layerHKB}
\end{figure}

The presentation of atomic knowledge can be various. Instead of utilizing declarative sentences or subject-relationship-object tuples, we propose using questions as knowledge indexes to further bridge the gap between stored knowledge and query. Unlike the semantic tagging process, in knowledge atomizing process, we input the document chunk to LLM as context, ask it to generate relevant questions that can be answered by the given chunk as many as possible. These generated atomic questions are saved as the \textit{atomic question} tags together with the given chunk. An example of knowledge atomizing is demonstrated in Figure~\ref{fig: atomic-example}(c), where the atomic questions encapsulate various aspects of the knowledge contained within the chunk.
A hierarchical knowledge base can accommodate queries of varying granularity. Figure \ref{fig: 2layerHKB} illustrates the retrieval process from an atomic knowledge base comprising chunks and atomic questions. Queries can directly retrieve reference chunks as usual. Additionally, since each chunk is tagged with multiple atomic questions, an atomic query can be used to locate relevant atomic questions, which then leads to the associated reference chunks.

\subsubsection{Knowledge-Aware Task Decomposition}

For a specific task, multiple decomposition strategies might be applicable. Consider Q2 in Figure~\ref{fig-qeustion-type} as an example. The two-step analytical reasoning process depicted may be effective if an interchangeable biosimilar products list is available. However, if only a general list of biosimilar products exists, with attributes dispersed throughout multiple documents, a different decomposition strategy may be necessary: (1) Retrieve the biosimilar product list; (2) Determine whether each product is interchangeable; (3) Count the total number of interchangeable products. The critical factor in selecting the most effective decomposition approach lies in understanding the contents of the specialized knowledge base.
Motivated by this, we design the \textit{Knowledge-Aware Task Decomposition} workflow, which is illustrated in Figure~\ref{fig: atomic-example}(a). The complete algorithm for task solving using Knowledge-Aware Task Decomposition is presented in Algorithm~\ref{alg:ka-decompose}.


\begin{algorithm}[ht]
\caption{Task Solving with Knowledge-Aware Decomposition}
\label{alg:ka-decompose}
\begin{algorithmic}[1]
    \State Initialize context $\mathcal{C}_0 \gets \pmb{\phi}$ 
    \For {$t=1,2,\ldots,N$}  \label{algline:loop_start}
        \State Generate atomic question proposals $\{\hat{q}_{i}^{t}\} \gets \mathcal{LLM}(q, \mathcal{C}_{t-1})$  \label{algline:gen_queries}
        \State\label{algline:retrieve_candidates} For each atomic question proposal $\hat{q}_{i}^{t}$, retrieve top-$K$ atomic candidates from knowledge base $$\{(q_{ij}^t, c_{ij}^t)\in\mathcal{KB}\mid\mathit{sim}(q_{ij}^t, \hat{q}_{i}^{t})\geq\delta\}$$  
        \State Select the most useful atomic question $q^t \gets \mathcal{LLM}(q, \mathcal{C}_{t-1}, \{q_{ij}^t\})$  \label{algline:select_atomic}
        \If{$q^t$ is $\mathit{None}$}
        \State $\mathcal{C}_t \gets \mathcal{C}_{t-1}$
        \State \textbf{break}
        \Else
        \State Fetch the relevant chunk $c^t$ corresponding to $q^t$
        \State Update context $\mathcal{C}_t \gets \mathcal{C}_{t-1} \cup \{c^t\}$
        \EndIf
    \EndFor
    \State Generate answer $\hat{a} \gets \mathcal{LLM}(q, \mathcal{C}_t)$ \label{algline:gen_answer}
\end{algorithmic}
\end{algorithm}

The reference context $\mathcal{C}_t$ is initialized as an empty set, and the original question is denoted by $q$. As illustrated in the for-loop starting at line~\ref{algline:loop_start} of the algorithm, in the $t$-th iteration, we use an LLM, denoted by $\mathcal{LLM}$, to generate query proposals potentially useful for task completion, denoted as ${\hat{q}_{i}^t}$. In this step, the chosen reference chunks $\mathcal{C}_t$ are provided as context to avoid generating proposals linked to already known knowledge. These proposals are then utilized as atomic queries to determine if relevant knowledge exists within the knowledge base. For each atomic question proposal, we retrieve its relevant atomic question candidates along with their source chunks $\{(q_{ij}^t, c_{ij}^t)\}$ from the knowledge base, denoted as $\mathcal{KB}$. We can use any score metric $\mathit{sim}$ to retrieve atomic questions. In our experiment, we use cosine similarity of their corresponding embeddings to retrieve all top-$K$ atomic questions, provided their similarity to a proposed atomic question is greater than or equal to a given threshold $\delta$. With the original question $q$, the accumulated context $\mathcal{C}_t$, and the list of retrieved atomic questions ${q_{ij}^t}$, $\mathcal{LLM}$ selects the most useful atomic question $q^t$ from ${q_{ij}^t}$ and retrieves the relevant chunk $c^t$. This retrieved chunk is aggregated into the reference context $\mathcal{C}_t$ for the next round of decomposition. Knowledge-aware decomposition can iterate up to $N$ times, where $N$ is a hyperparameter set to control computational cost. The iteration process can be terminated early if there are no high-quality question proposals, no highly relevant atomic candidates retrieved, no suitable atomic knowledge selections, or if the $\mathcal{LLM}$ determines that the acquired knowledge is sufficient to complete the task. Finally, the accumulated context $\mathcal{C}_t$ is utilized to generate answer $\hat{a}$ for the given question $q$ in line~\ref{algline:gen_answer}.


\subsubsection{Knowledge-Aware Task Decomposer Training}

\begin{figure}[t]
    \begin{center}
        \includegraphics[width=0.8\linewidth]{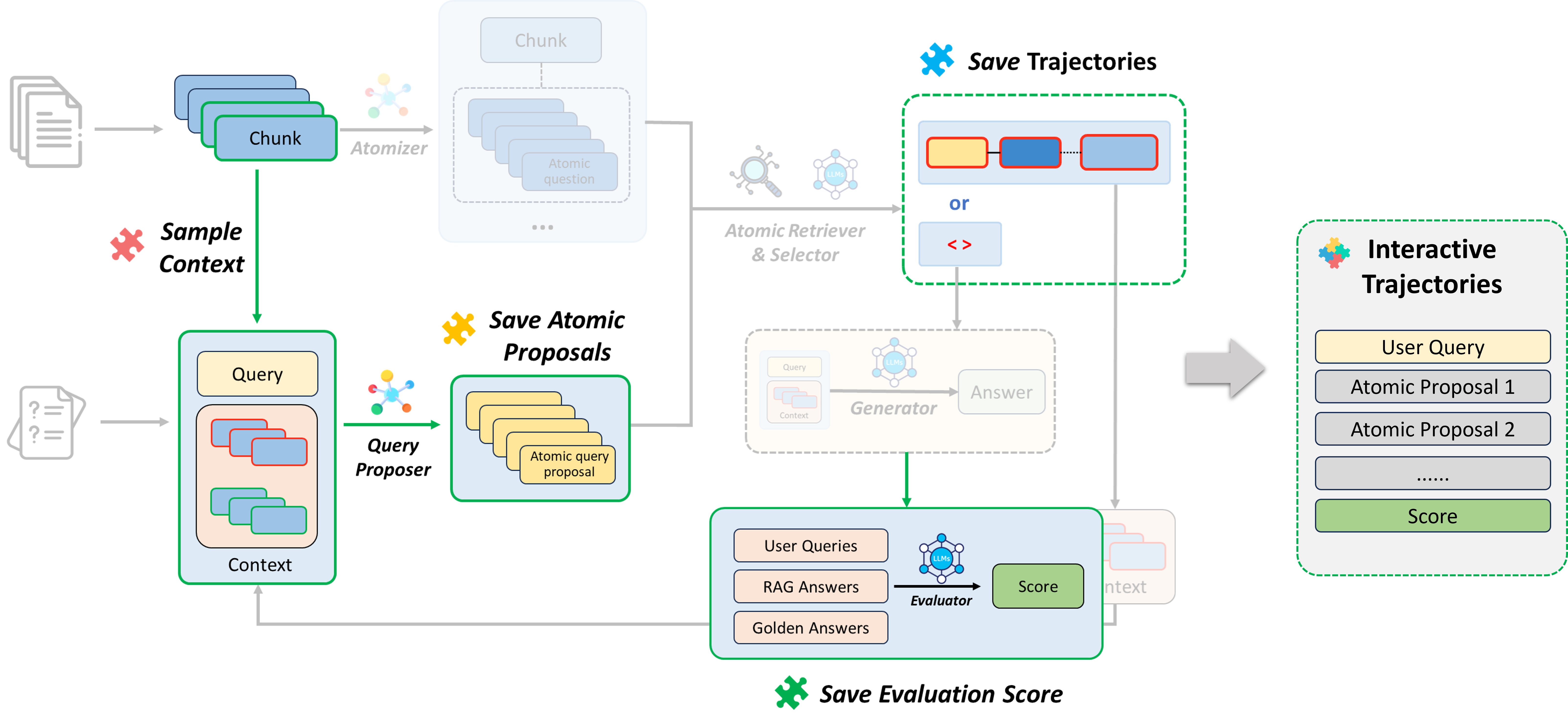}
    \end{center}
    \vspace{-2mm}
    \caption{ Data collection process for decomposer training, comprising four main components: a) sampling data chunks from the context sampling pool to serve as the reference context for question decomposition, b) saving the generated atomic query proposals, c) after retrieval and selection, saving the chosen atomic query proposals as part of the reasoning trajectories, d) evaluating the answer to generate a score.} 
    \label{fig: data_collection}
\end{figure}

\begin{algorithm}[ht]
\caption{Collect Data to Train a Knowledge-Aware Decomposer}
\label{alg:ka-decompose-data}
\begin{algorithmic}[1]
    \State Initialize context $\mathcal{C}_0 \gets \pmb{\phi}$ 
    \State Initialize a dictionary to store scores of each chunk $\mathcal{S}=\{c: 0\mid\forall c\}$
    \State Initialize a dictionary to store visits of each chunk $\mathcal{V}=\{c: 1\mid\forall c\}$
    \For {$t=1,2,\ldots,N$}
        \State $c_{\mathit{sampled}} = \mathit{argmax}_{c}(\mathcal{S}(c)+\alpha\sqrt{\frac{\ln t}{\mathcal{V}(c)}})$
        \State Generate atomic question proposals $\{\hat{q}_{i}^{t}\} \gets \mathcal{LLM}(q, \mathcal{C}_{t-1}\cup\{c_{\mathit{sampled}}\})$  
        \State For each $\hat{q}_{i}^{t}$, retrieve top-$K'$ atomic candidates from knowledge base $$\textbf{AC}_t=\{(q_{ij}^t, c_{ij}^t)\in\mathcal{KB}\mid\exists\hat{q}_{i}^{t}.\mathit{sim}(q_{ij}^t, \hat{q}_{i}^{t})\geq\delta'\}$$
        \State Initialize a list $\textbf{RAP}_t=\pmb{\phi}$ to store the most relevant atomic questions
        \For {$(q, c)\in \textbf{AC}_t$}
            \If{$\exists\hat{q}_{i}^{t}.\mathit{sim}(q, \hat{q}_{i}^{t})\geq\delta$}
                \State $\textbf{RAP}_t\xleftarrow{} q$
            \Else
                \State $\mathcal{S}(c) = \mathcal{S}(c)+\max\{\mathit{sim}(q, \hat{q}_{i}^{t})\mid\forall \hat{q}_{i}^{t}\}$
            \EndIf
        \EndFor
        \State Select the most relevant atomic question $q^t \gets \mathcal{LLM}(q, \mathcal{C}_{t-1}, \textbf{RAP}_t)$  
        \If{$q^t$ is $\mathit{None}$}
        \State $\mathcal{C}_t \gets \mathcal{C}_{t-1}$
        \State \textbf{break}
        \Else
        \State Fetch the relevant chunk $c^t$ corresponding to $q^t$
        \State Update context $\mathcal{C}_t \gets \mathcal{C}_{t-1} \cup \{c^t\}$
        \State Update score of $c_t$ $\mathcal{S}(c_t)=0$
        \State Increase visit counts of $c_t$ $\mathcal{V}(c_t)=\mathcal{V}(c_t)+1$
        \EndIf
    \EndFor
    \State Generate answer $\hat{a} \gets \mathcal{LLM}(q, \mathcal{C}_t)$
\end{algorithmic}
\end{algorithm}

\begin{figure}[t]
    \begin{center}
        \includegraphics[width=0.98\linewidth]{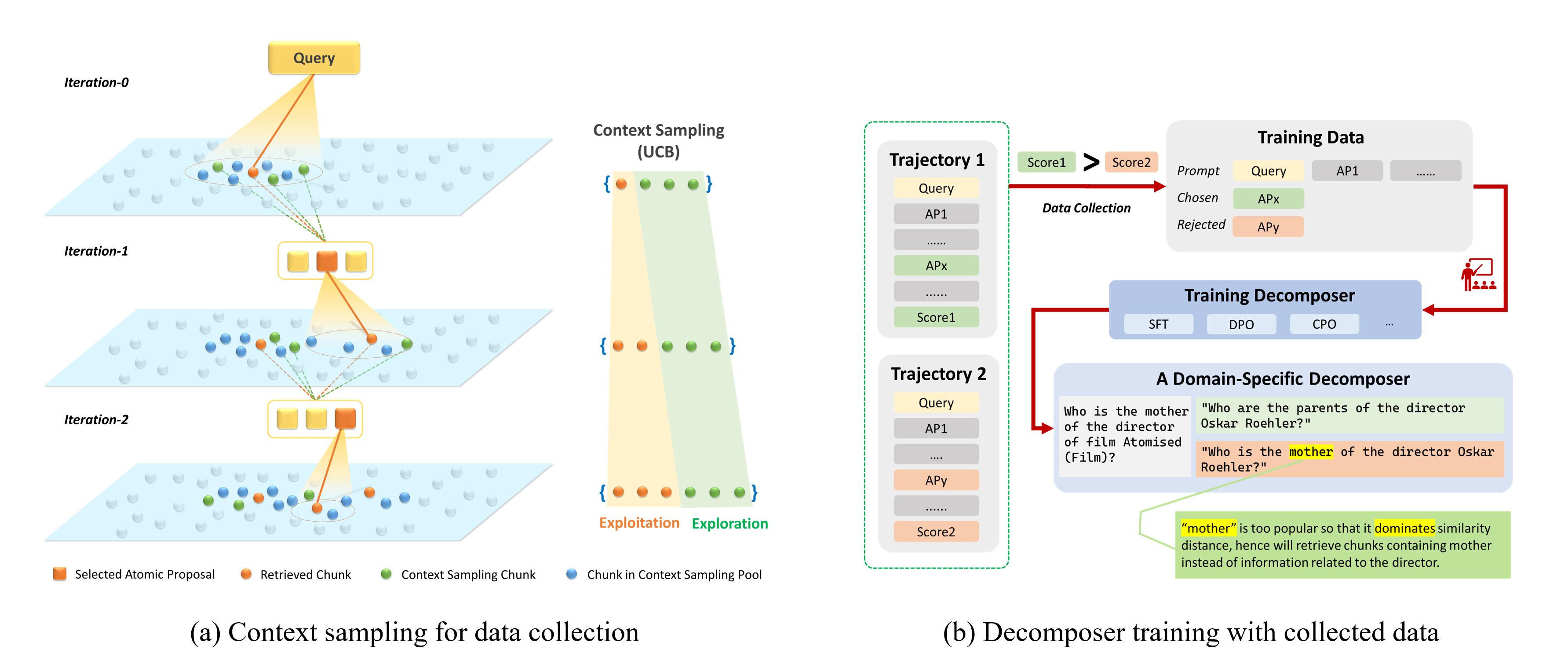}
    \end{center}
    \vspace{-2mm}
    \caption{ An example of context sampling and an illustration of decomposer training with collected data.} 
    \label{fig: decomposer_train}
    \vspace{-2mm}
\end{figure}

It is worth mentioning that knowledge-aware decomposition can be a learnable component. This trained proposer can then directly suggest atomic queries $q^t$ during inference, which means lines~\ref{algline:gen_queries} to~\ref{algline:select_atomic} in Algorithm~\ref{alg:ka-decompose} can be replaced by a single call to this learned proposer, thereby reducing both inference time and computational cost. In order to train the knowledge-aware decomposer, we collect data about the rationale behind each step by sampling context and creating diverse interaction trajectories. With this data collected, we train a decomposer that can incorporate domain-specific rationale into the task decomposition and result-seeking process.

The data collection process, as depicted in Figure~\ref{fig: data_collection} and Algo.~\ref{alg:ka-decompose-data}, implements a sophisticated dual-dictionary system for managing and tracking information. Our system utilizes two primary data structures: dictionary $\mathcal{S}$ for maintaining comprehensive score records, and dictionary  $\mathcal{V}$ for systematically tracking visit frequencies of candidate chunks. During the initialization phase, we establish baseline values by setting all scores to zero and initializing visit counters to one, creating a foundation for dynamic updates throughout the subsequent processing stages. 

In each iteration of our decomposition process, the system executes a detailed retrieval operation targeting the top-$K'$ chunks demonstrating maximum relevance to the current atomic question. These chunks must satisfy our similarity threshold criterion (specifically, similarity exceeding $\delta'$, where  $\delta'<\delta$), with $K'$ intentionally configured to be larger than $K$ to ensure comprehensive coverage. Following this initial retrieval, we carefully select and integrate the data chunks corresponding to the top-$K$ most relevant atomic retrieved pairs into the context. For those retrieved chunks that do not make it into the top-$K$ selection, we systematically incorporate them into $\mathcal{S}$ and methodically update their scores based on precisely calculated relevance metrics.

To ensure comprehensive exploration of the solution space, we have implemented an advanced sampling mechanism that intelligently selects additional chunks from $\mathcal{S}$ when available, incorporating them seamlessly into the reference context. Our implementation leverages the Upper Confidence Bound~\cite{auer2002using} (UCB) algorithm for context sampling, establishing a balanced approach between exploitation and exploration. The exploitation component manifests through the retriever-selected chunks, focusing on options with currently highest estimated rewards to optimize immediate performance gains. Conversely, the exploration aspect is fulfilled through context sampling from $\mathcal{S}$, enabling the systematic investigation of less-certain options to accumulate valuable data and potentially uncover superior long-term alternatives.

This meticulously crafted strategy serves a dual purpose: it not only facilitates the generation of diverse and comprehensive atomic query proposals but also enables systematic exploration of multiple potential reasoning pathways. Through this sophisticated approach, we progressively work toward deriving optimal final answers while maintaining a balance between immediate performance optimization and long-term discovery of potentially superior solutions.

We record atomic proposals (AP), interactive trajectories, and answer scores to support decomposer training. 
For each specialized domain, interactive trajectories featuring distinct reasoning paths are gathered for decomposer training.  This allows us to use the answer score as a supervised signal to train the decomposer. The decomposer training process is depicted in Figure~\ref{fig: decomposer_train}. By incorporating preferences in the form of answer scores, the decomposer training can capture domain-specific decomposition rules, thereby adapting the decomposer to meet domain requirements. 

Looking ahead, there are several promising avenues for implementing and enhancing our proposed decomposer. We could leverage well-established algorithms such as supervised fine-tuning (SFT) and direct policy optimization (DPO)~\cite{rafailov2023direct} to train an effective decomposer based on existing LLMs. The practical implementation and performance evaluation of this comprehensive procedure, including detailed empirical analysis and comparative studies, will be addressed in future research work to thoroughly demonstrate its effectiveness and potential applications.

\subsection{Level-3: Predictive Question focused RAG System}
\begin{figure}[t]
	\begin{center}
		\includegraphics[width=0.75\linewidth]{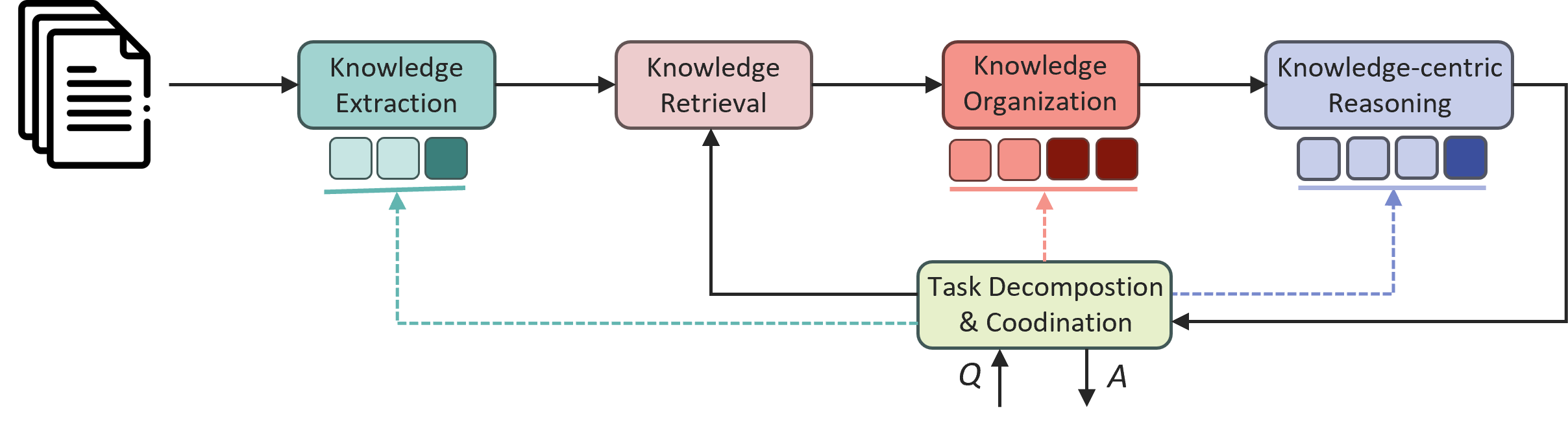}
	\end{center}
	\vspace{-2mm}
	\caption{Overview of L3-RAG framework. The squares $($\includegraphics[height=0.2cm]{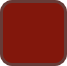}$)$ indicate knowledge structuring and knowledge induction in knowledge organization module, while the square $($\includegraphics[height=0.2cm]{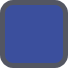}$)$ represents forecasting sub-module in knowledge-centric reasoning module.} 
	\label{fig-L3}
\end{figure}

In the L3 system, there is an increased emphasis on knowledge-based prediction capability, which necessitates effective knowledge collection, organization, and the construction of forecasting rationale. To address this, we leverage the task decomposition and coordination module to build forecasting rationale based on the organized knowledge, which is collected and organized from the retrieved knowledge. The framework of L3 system is illustrated in Figure\ref{fig-L3}. To ensure the retrieved knowledge is well-prepared for advanced analysis and forecasting, the knowledge organization module is enhanced with specialized submodules dedicated to the structuring and organization of knowledge. These submodules streamline the process of transforming raw retrieved knowledge into a structured, coherent format, optimizing it for subsequent reasoning and predictive tasks. 
For example, in the FDA scenario referred in Figure~\ref{fig-qeustion-type}, data from multiple sources—such as medicine labels, clinical trials, and application forms—are integrated into the multi-layer knowledge base. The knowledge structuring submodule follows the instruction from task decomposition module to collect and organize the relevant knowledge (e.g. medicine names with their approval dates) retrieved from knowledge base. The knowledge induction submodule further categorizes this structured knowledge, such as by approval date, to facilitate further statistics analysis and prediction.


Given the limitations of LLMs in applying specialized reasoning logic, their effectiveness in predictive tasks can be restricted. To overcome this, the knowledge-centric reasoning module is enhanced with a forecasting submodule, enabling the system to infer outcomes based on the input queries and the organized knowledge (e.g. total numbers of medicines approved per year). This forecasting submodule allows the system to not only generate answers based on historical knowledge, but also make projections, providing a more robust and dynamic response to complex queries. By integrating advanced knowledge structuring and prediction capabilities, the L3 system can manage and utilize a more complex and dynamic knowledge base effectively.  







\subsection{Level-4: Creative Question focused RAG System}
\begin{figure}[H]
	\begin{center}
		\includegraphics[width=0.9\linewidth]{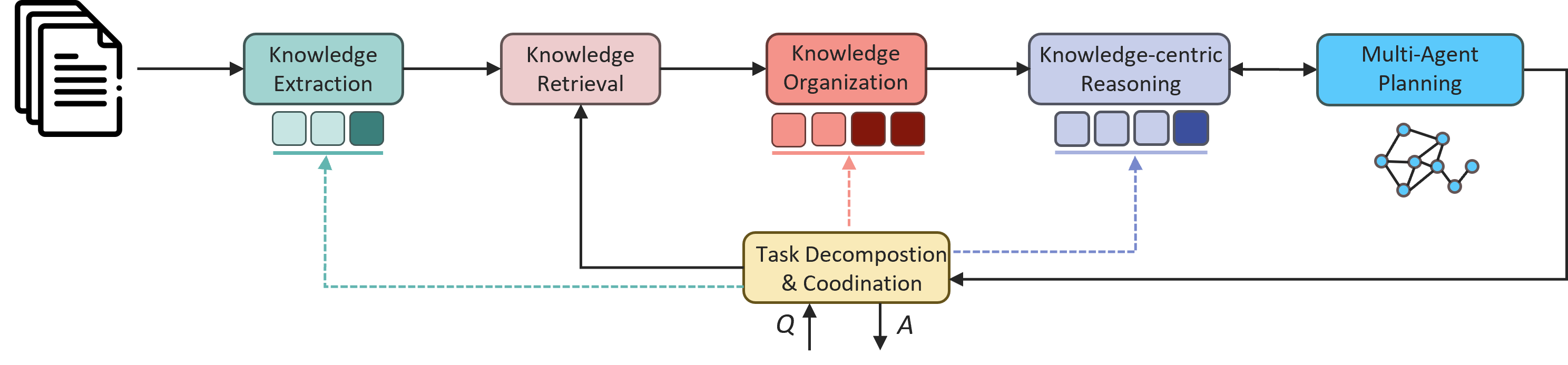}
	\end{center}
	\vspace{-2mm}
	\caption{Overview of L4-RAG framework. The multi-agent planning module is introduced to enable multi-perspective thinking.} 
	\label{fig-L4}
\end{figure}

The L4 system implementation is characterized by the integration of multi-agent systems to facilitate multi-perspective thinking. Addressing creative questions requires creative thinking that draws on factual information and an understanding of underlying principles and rules. At this advanced level, the primary challenges include extracting coherent logical rationales from a retrieved knowledge, navigating complex reasoning processes with numerous influencing factors, and assessing the quality of responses to creative, open-ended questions. 
To tackle these challenges, the system coordinates multiple agents, each contributing unique insights and reasoning strategies, as illustrated in Figure\ref{fig-L4}. These agents operate in parallel, synthesizing various thought processes to generate comprehensive and coherent solutions. This multi-agent architecture supports the parallel processing and integration of diverse reasoning paths, ensuring effective management and response to intricate queries.
By simulating diverse viewpoints, the L4 system enhances its ability to tackle creative questions, generating innovative ideas rather than predefined solutions. The coordinated outputs from multiple agents not only enrich the reasoning process but also provide users with comprehensive perspectives, fostering creative thinking and inspiring novel solutions to complex problems.

\section{Evaluation and Metrics}

To validate the effectiveness of our proposed method, we conduct experiments on both open-domain benchmarks and domain-specific benchmarks. We delineate the evaluation metrics and methods employed to assess the performance of proposed knowledge-aware task decomposition method in Section~\ref{subsec:exp_setup}. The evaluation results on three open-domain benchmarks are presented in Section~\ref{subsec:open_evaluation}, while the results on two legal domain-specific benchmarks in Section~\ref{subsec:legal_evaluation}. Furthermore, we present in-depth analysis through three real case studies in Section~\ref{subsec:real_cases}, which highlight the superiority of our method compared to existing decomposition approaches.

\subsection{Experimental Setup}  \label{subsec:exp_setup}

\paragraph{Methods} To thoroughly evaluate the performance of our proposed knowledge-aware decomposition approach (described in Section \ref{subsec:l2_atomic_decompo}), we have selected a variety of baseline methods that represent different strategies for task-solving with LLMs. We include \textbf{Zero-Shot CoT}\cite{kojima2022zeroshotcot} to assess the inherent reasoning capabilities and embedded knowledge of the underlying LLM without additional context. \textbf{Naive RAG}\cite{lewis2020retrieval}, which introduces external knowledge through retrieval, serves as a benchmark for evaluating the incremental benefits of augmented knowledge. The \textbf{Self-Ask} framework\cite{press2023measuringnarrowingcompositionalitygap} is employed to investigate the impact of an iterative question decomposition and answering strategy on task performance. Additionally, \textbf{GraphRAG}~\cite{edge2024localglobalgraphrag} is evaluated in both local and global modes to assess the impact of knowledge graph-based methods on multi-hop reasoning tasks.

To ensure a fair comparison and to highlight the influence of hierarchical knowledge structures, we have extended Naive RAG and Self-Ask to utilize both a general flat knowledge base, denoted as \textit{R}, and a hierarchical retriever, denoted as \textit{H-R}, as introduced in Figure~\ref{fig: 2layerHKB}. The hierarchical retriever (\textit{H-R}) utilizes the questions or follow-up questions to retrieve chunks through both path (a) and path (b) at the same time. The retrieved chunks from both paths are then aggregated to form a comprehensive reference context for LLM to answer each question, potentially enhancing the relevance of the provided context.

The experimental methods are summarized as follows:

\begin{itemize}
    \item \textbf{Zero-Shot CoT}: Questions are addressed using solely the Chain-Of-Thought (CoT) technique, which prompts the LLMs to articulate its reasoning process step-by-step without the aid of example demonstrations or supplemental context. This method assesses the LLMs' intrinsic knowledge and reasoning capabilities in a zero-shot setting.
    
    \item \textbf{Naive RAG w/ R}: This approach employs dense retrieval from a flat knowledge base to procure relevant information for each question. The knowledge base consists of pre-embedded chunks are matched to the original question based on semantic similarity. The retrieval process is direct, without any intermediate task decomposition.
    
    \item \textbf{Naive RAG w/ H-R}: This method extends the Naive RAG framework by incorporating a hierarchical retrieval process (\textit{H-R}) that operates through two concurrent paths. Path (a) performs a direct retrieval of knowledge chunks in response to the original question, similar to the flat retrieval approach. Path (b), on the other hand, use the original question again to find the relevant atomic questions and obtain the corresponding chunks. The combined output from both paths is then aggregated, creating a rich reference context.

    \item \textbf{Self-Ask}: This method employs a task decomposition strategy wherein the LLMs is prompted to iteratively generate and answer follow-up questions, thereby breaking down complex problems into more manageable sub-tasks. General demonstrations illustrating the logic and methodology of task decomposition are provided for all benchmarks to guide the LLMs' reasoning process. As detailed in the original paper ~\cite{press2023measuringnarrowingcompositionalitygap}, the framework encourages the LLMs to engage in a recursive dialogue with itself, generating intermediate answers that progressively build towards the final answer. In this setting, the LLMs relies solely on its inherent knowledge base, as no external contexts are introduced to aid in answering the follow-up questions.

    \item \textbf{Self-Ask w/ R}: Building upon the \textbf{Self-Ask} method, this setting introduces an additional retrieval component, for each follow-up question generated by the LLMs, relevant chunks are retrieved from a flat knowledge base to provide a reference context. The retrieval process uses the follow-up question as the query. This approach seeks to combine the benefits of iterative task decomposition with rich external knowledge from retrieval, potentially improving the LLMs' performance on complex reasoning tasks.
    
    \item \textbf{Self-Ask w/ H-R}: This variant of the Self-Ask method enhances the retrieval process by utilizing a hierarchical knowledge base, as opposed to the flat one used in \textbf{Self-Ask w/ R}. When the LLMs generates follow-up questions, these are employed as queries in a dual-path retrieval system, specifically paths (a) and (b) in Figure~\ref{fig: 2layerHKB}. The outputs from both retrieval paths are then aggregated to form a richer reference context.

    \item \textbf{GraphRAG Local}: In this approach, the flat knowledge base is pre-processed to construct a knowledge graph in accordance with the public guidance. The inference is run in local mode.

    \item \textbf{GraphRAG Global}: The inference is run in global mode in this setting.
        
    \item \textbf{Ours}: The proposed knowledge-aware decomposition method iteratively decomposes complex questions into sub-questions and retrieves relevant knowledge up to a maximum of five iterations. This process limits the context for the final answer to the five most useful knowledge chunks.
\end{itemize}

\paragraph{Metrics} For maintaining consistency with established benchmarks, two conventional metrics are adopted in our experimental evaluation: \textbf{Exact Match (EM)}, which assesses whether the response is identical to a predefined correct answer, and the \textbf{F1} score, which is the harmonic mean of precision and recall at the token level.
During evaluation, we noticed that the LLM sometimes produced responses more verbose than expected, even when the QA prompt aimed to limit output style.
To more accurately gauge the responses' alignment with the intended answers—beyond mere lexical matching—we introduced a novel evaluation metric employing \textit{GPT-4}. In this process, \textit{GPT-4} acts as an evaluator, assessing the correctness of a response in relation to the question and the correct answer labels. We refer to this metric as \textbf{Accuracy (Acc)}. Upon manual inspection of a sample set, the judgments rendered by \textit{GPT-4} demonstrate complete agreement with human evaluators, affirming the reliability of this metric.

Furthermore, we encountered situations where a method achieves high accuracy (Acc) scores yet registers low F1 scores. To elucidate the underlying factors of such discrepancies, we also report on the \textbf{Recall} and \textbf{Precision} of the generated responses. Recall measures the proportion of relevant tokens from the answer labels that are captured in the response, while precision evaluates the relevance of the tokens in the generated answer with respect to the correct labels.
Specifically, in cases where multiple correct answer labels are available, we employ a conservative scoring approach for \textbf{EM}, \textbf{F1}, \textbf{Precision}, and \textbf{Recall} by retaining the highest score achieved. This approach is designed to equitably consider the range of correct answers that the LLM may generate. It should be noted that, in the context of computing \textbf{Accuracy (Acc)}, all admissible answer labels are furnished concurrently to the evaluation process, resulting in a singular Accuracy score.

The metrics employed in this evaluation — \textbf{Exact Match (EM)}, \textbf{F1}, \textbf{Precision}, \textbf{Recall}, and \textbf{Accuracy (Acc)} — are primarily suited for questions categorized as L1 and L2, which are characterized by their association with ground truth answers that are factual and definitive. However, the utility of these metrics diminishes for predictive and creative questions, namely the L3 and L4 questions, where answers are inherently uncertain or subjective, and no single correct response exists. For L3 questions, alternative assessment methods such as trend judgment and qualitative analysis become more appropriate to capture the predictive validity of the responses. Furthermore, for L4 questions, which demand a higher degree of insight or innovation, it is essential to evaluate answers through a multi-faceted lens, considering criteria such as relevance, diversity, comprehensiveness, uniqueness, and inspiration to fully appreciate the depth and originality of the approaches' responses.

\paragraph{LLM and Hyper-parameters} In our experiments, we employ GPT-4 (1106-Preview version) across all the methods outlined previously. For the knowledge extraction phase, we utilize a \textit{temperature} setting of $0.7$ specifically for the \textit{Knowledge Atomizing} process, promoting a balance between diversity and determinism in the generated atomic knowledge. Conversely, during all question-answering (QA) steps in each method, we implement a \textit{temperature} of $0$, ensuring consistent responses from the model.
Regarding the retrieval component, we engage the \textit{text-embedding-ada-002} (version 2) as our embedding model for both the general flat knowledge bases and the hierarchical knowledge bases. For the general flat knowledge bases, the retriever is configured to fetch up to 16 knowledge chunks, applying a retrieval score threshold of $0.2$. In the case of hierarchical knowledge bases, the retriever is initially set to retrieve a maximum of 8 chunks with a more stringent threshold of $0.5$. Subsequently, an additional 4 chunks can be retrieved via each atomic query posed.

\subsection{Evaluation on Open-Domain Benchmarks}  \label{subsec:open_evaluation}

In this subsection, we demonstrate the performance of our method across three open-domain benchmarks. To ensure a fair and objective evaluation, particularly in the context of real-world industrial applications, we have selected three widely-recognized multi-hop datasets: HotpotQA~\cite{yang2018hotpotqa}, 2WikiMultiHopQA~\cite{ho2020twowiki}, and MuSiQue~\cite{trivedi2022musique}. Below, we provide a brief overview of these datasets, noting that our method does not leverage the question type information nor the number of hops information during the solving process, as our approach is designed to be agnostic to such classifications.

\paragraph{HotpotQA} The HotpotQA dataset is a well-known multi-hop QA benchmark primarily consisting of 2-hop questions, each associated with 10 Wikipedia paragraphs. Among these, some paragraphs contain supporting facts essential to answering the question, while the rest serve as distractors. The dataset also includes a \textit{question type} field, which delineates the logical reasoning required—\textit{comparison} questions involve contrasting two entities, and \textit{bridge} questions require inferring the bridge entity, or inferring the property of an entity through an intermediary entity, or locating the answer entity~\cite{yang2018hotpotqa}. The \textit{comparison} questions in HotpotQA align with the \textit{comparative} questions defined in Section~\ref{subsec:task_classification}. Similarly, \textit{bridge} questions correspond to either \textit{bridging} questions or \textit{summarizing} questions, depending on the complexity of the rationale required. Although our method operates independently of these types, their description here exemplifies the nature of questions within the dataset and contextualizes the expected performance variance across different benchmarks.

\paragraph{2WikiMultiHopQA} Inspired by HotpotQA, 2WikiMultiHopQA expands the diversity of question types. It retains the \textit{comparison} type from HotpotQA and introduces \textit{inference} and \textit{compositional} questions that evolve from the \textit{bridge} type by focusing on entity attribute deduction and entity location, respectively. Additionally, the \textit{bridge comparison} type is a novel category that requires a synthesis of \textit{bridge} and \textit{comparison} reasoning. In this dataset, the \textit{comparison} questions correspond to the \textit{comparative} questions defined in Section~\ref{subsec:task_classification}, akin to those in HotpotQA. The \textit{inference} questions are analogous to \textit{bridging} questions, and the \textit{compositional} questions are similar to \textit{summarizing} questions as described in the same section. The \textit{bridge\_comparison} questions, due to their hybrid nature and increased complexity, also fall under the \textit{summarizing} questions category. This dataset typically presents 2-hop to 4-hop questions, each accompanied by 10 Wikipedia paragraphs containing supporting facts and distractors. While these types inform the dataset's structure, they are not utilized by our method, which treats all questions uniformly regardless of their categorization.

\paragraph{MuSiQue} Addressing the issue that many multi-hop questions can be solved via shortcuts—arriving at correct answers without proper reasoning—MuSiQue implements stringent filters and additional mechanisms specifically designed to encourage connected reasoning, as reported by Trivedi et al.~\cite{trivedi2022musique}. 
Unlike the other datasets, MuSiQue does not categorize questions by type, but it does provide explicit information on the number of hops required for each question, ranging from 2 to 4 hops. Each question is associated with 20 context paragraphs, which introduce a mix of relevant and irrelevant information, further complicating the task of discerning the correct reasoning path. This explicit hop information, while not used by our method, underscores the complexity of the dataset and the robustness required by models to handle such challenges effectively.

In our experiments, we randomly sample 500 QA data from the \textit{dev} set of each dataset, without consideration for question type nor number of hops, to ensure randomness. We compile the context paragraphs from all sampled QA data into a single knowledge base for each benchmark, creating a more complex retrieval scenario. This design choice is aimed at rigorously assessing our model's question decomposition and relevant context retrieval abilities. Table \ref{tab:q_type_sum} outlines the distribution of question types within our sampled sets, offering insight into the variety of reasoning challenges presented in our evaluation, though this does not directly impact our method.


\begin{table}[t]
\fontsize{8}{9}\selectfont
\renewcommand{\arraystretch}{1.1}
    \centering
    \caption{Distribution of question types across three multi-hop QA datasets.}
    \vspace{5mm}
    \begin{minipage}[t][1.6cm][t]{0.3\textwidth}
        \centering
        \begin{subtable}[b]{\textwidth}
            \centering
            \begin{tabular}{c|cc}
                \hline
                Type & Count & Ratio \\
                \hline
                comparison & 107 & 21.4\% \\
                bridge & 393 & 78.6\% \\
                \hline
            \end{tabular}
        \caption{HotPotQA}    
        \end{subtable}
    \end{minipage}
    \hfill
    \begin{minipage}[t][1.6cm][t]{0.35\textwidth}
        \centering
        \begin{subtable}[b]{\textwidth}
            \centering
            \begin{tabular}{c|cc}
                \hline
                Type & Count & Ratio \\
                \hline
                comparison & 132 & 26.4\% \\
                inference & 64 & 12.8\% \\
                compositional & 196 & 39.2\% \\
                bridge\_comparison & 108 & 21.6\% \\
                \hline
            \end{tabular}
        \caption{2WikiMultiHopQA}    
        \end{subtable}
    \end{minipage}
    \hfill
    \begin{minipage}[t][1.6cm][t]{0.3\textwidth}
        \centering
        \begin{subtable}[b]{\textwidth}
            \centering
            \begin{tabular}{c|cc}
                \hline
                \#Hops & Count & Ratio \\
                \hline
                2 & 263 & 52.6\% \\
                3 & 169 & 33.8\% \\
                4 & 68 & 13.6\% \\
                \hline
            \end{tabular}
            \caption{MuSiQue}        
        \end{subtable}
    \end{minipage}
    \label{tab:q_type_sum}
\end{table}

\begin{table}[t]
\fontsize{9}{11}\selectfont
\renewcommand{\arraystretch}{1.1}
\centering
\caption{Performance comparison on HotpotQA. Best in bold, second-best underlined.}
\vspace{3mm}
\begin{tabular}{l|ccccc}
\hline
Method  & EM  & F1  & Acc  & Precision  & Recall \\
\hline
Zero-Shot CoT  & 32.60  & 43.94  & 53.60  & 46.56  & 43.97 \\
Naive RAG w/ R  & \underline{56.80}  & \underline{72.67}  & 82.60  & \underline{74.52}  & 74.86 \\
Naive RAG w/ H-R  & 54.80  & 70.25  & 81.60  & 72.56  & 72.24 \\
Self-Ask  & 28.80  & 43.61  & 59.60  & 43.49  & 56.21 \\
Self-Ask w/ R  & 44.80  & 63.08  & 81.00  & 63.23  & 74.57 \\
Self-Ask w/ H-R  & 47.20  & 64.24  & 82.20  & 64.27  & 75.95 \\
GraphRAG Local  & 0.00  & 10.66  & \textbf{89.00}  & 5.90  & \textbf{83.07} \\
GraphRAG Global  & 0.00  & 7.42  & 64.80  & 4.08  & 63.16 \\
Ours  & \textbf{61.20}  & \textbf{76.26}  & \underline{87.60}  & \textbf{78.10}  & \underline{78.95} \\
\hline
\end{tabular}
\label{tab:hotpotqa_results}
\end{table}

\begin{table}[t]
\fontsize{9}{11}\selectfont
\renewcommand{\arraystretch}{1.1}
\centering
\caption{Performance comparison on 2WikiMultiHopQA. Best in bold, second-best underlined.}
\vspace{3mm}
\begin{tabular}{l|ccccc}
\hline
Method  & EM  & F1  & Acc  & Precision  & Recall \\
\hline
Zero-Shot CoT  & 35.67  & 41.40  & 43.87  & 41.43  & 43.11 \\
Naive RAG w/ R  & 51.20  & 59.74  & 62.80  & 59.06  & 62.30 \\
Naive RAG w/ H-R  & \underline{51.40}  & 59.73  & 63.00  & 59.36  & 62.43 \\
Self-Ask  & 23.80  & 37.49  & 51.60  & 34.56  & 60.72 \\
Self-Ask w/ R  & 46.80  & \underline{64.17}  & 79.80  & 61.17  & \textbf{80.21} \\
Self-Ask w/ H-R  & 48.00  & 63.99  & \underline{80.00}  & \underline{61.30}  & \underline{79.56} \\
GraphRAG Local  & 0.00  & 11.83  & 71.20  & 6.74  & 75.17 \\
GraphRAG Global  & 0.00  & 7.35  & 45.00  & 4.09  & 55.43 \\
Ours  & \textbf{66.80}  & \textbf{75.19}  & \textbf{82.00}  & \textbf{74.04}  & 78.87 \\
\hline
\end{tabular}
\label{tab:2wiki_results}
\end{table}

\begin{table}[ht]
\fontsize{9}{11}\selectfont
\renewcommand{\arraystretch}{1.1}
\centering
\caption{Performance comparison on MuSiQue. Best in bold, second-best underlined.}
\vspace{3mm}
\begin{tabular}{l|ccccc}
\hline
Method  & EM  & F1  & Acc  & Precision  & Recall \\
\hline
Zero-Shot CoT  & 12.93  & 22.90  & 23.47  & 24.40  & 24.10 \\
Naive RAG w/ R  & \underline{32.00}  & 43.31  & 44.40  & \underline{44.42}  & 47.29 \\
Naive RAG w/ H-R  & 30.40  & 41.30  & 43.40  & 42.06  & 44.53 \\
Self-Ask  & 16.40  & 27.27  & 35.40  & 26.33  & 37.65 \\
Self-Ask w/ R  & 28.40  & 42.54  & 49.80  & 41.13  & 53.37 \\
Self-Ask w/ H-R  & 29.80  & \underline{44.05}  & \underline{54.00}  & 42.47  & \underline{55.89} \\
GraphRAG Local  & 0.60  & 9.62  & 49.80  & 5.73  & 55.82 \\
GraphRAG Global  & 0.00  & 5.16  & 44.60  & 2.82  & 52.19 \\
Ours  & \textbf{46.40}  & \textbf{56.62}  & \textbf{59.60}  & \textbf{57.45}  & \textbf{59.53} \\
\hline
\end{tabular}
\label{tab:musique_results}
\end{table}

\paragraph{Overall Performance} The evaluation results across HotpotQA, 2WikiMultiHopQA, and MuSiQue are presented in Table \ref{tab:hotpotqa_results}, Table \ref{tab:2wiki_results}, and Table \ref{tab:musique_results}, respectively. 
If we hypothesize that the highest achievable performance on each benchmark may reflect its relative difficulty, a tentative ranking from easiest to most challenging would be: HotpotQA, 2WikiMultiHopQA, and MuSiQue.
Our observations suggest that for HotpotQA, considered the least challenging, the GraphRAG in local mode and our method are closely competitive, with minor performance disparities. However, as the difficulty increases for 2WikiMultiHopQA and MuSiQue, our method outperforms others.

The inclusion of retrieved context significantly enhances accuracy, with gains ranging from approximately 10\% (comparing Zero-Shot CoT and Naive RAG on MuSiQue) to around 29\% (on HotpotQA). This indicates that for simpler benchmarks, RAG equipped with naive knowledge retrieval could address simple multihop questions, leading to a significant accuracy boost. However, for more challenging benchmarks involving complex multihop questions, the accuracy improvement from naive knowledge retrieval is limited, underscoring the constrained reasoning capabilities of the LLMs.
By incorporating decomposition mechanisms, Self-Ask significantly enhances accuracy, especially on more challenging benchmarks. The combination of knowledge retrieval and Self-Ask decomposition yields superior results on 2WikiMultiHopQA and MuSiQue, compared to using a single mechanism. However, in the case of HotpotQA, all methods employing retrieval (except for GraphRAG in Global mode, which will be discussed later) attain accuracies above 80\%, with negligible differences between them.

Interestingly, the application of a hierarchical atomic knowledge base does not significantly impact Naive RAG's performance compared to Naive RAG with general flat knowledge base, potentially due to the embedding distance between the original multi-hop questions and the atomic questions of relevant contexts. Nonetheless, when combined with task decomposition, a hierarchical knowledge base shows more promise, as evidenced by the performance boost observed in Self-Ask with Hierarchical Retrieval (Self-Ask w/ H-R) compared to Self-Ask with Retrieval (Self-Ask w/ R), particularly on MuSiQue, which requires more complex reasoning. This improvement underscores the potential of hierarchical knowledge bases in enhancing the effectiveness of decomposition mechanisms in complex reasoning tasks.

Our proposed method focuses on knowledge-aware task decomposition, which performs decomposition with an awareness of available knowledge, effectively leveraging the atomic information provided by the hierarchical knowledge base. Experimental results demonstrate that our approach consistently outperforms other methods, validating its effectiveness in complex reasoning scenarios.

Regarding GraphRAG, originally designed for the query-focused summarization (QFS) task as outlined by~\cite{edge2024localglobalgraphrag}, we observe its suboptimal performance in both local and global modes compared to our method. Notably, GraphRAG exhibits a curious trend: it achieves higher accuracy and recall scores while performing lower on EM, F1, and Precision metrics.
A closer analysis of GraphRAG's outputs reveals a tendency to echo the query and include meta-information about the answer within its graph structure. Despite attempts to refine its QA prompt, this behavior persists. An illustrative example is presented in Table~\ref{tab:graphrag_example}, which shows GraphRAG Local's response to a question from HotpotQA.

\begin{table}[t]
\fontsize{9}{11}\selectfont
\renewcommand{\arraystretch}{1.1}
\centering
\caption{An Example of GraphRAG Local output on a HotpotQA question. The table showcases the tendency to repeat the question and include meta-information in its response.}
\vspace{3mm}
\begin{tabular}{l|p{3.6in}}
    \hline
    Question & Which country is home to Alsa Mall and Spencer Plaza? \\
    \hline
    Answer Labels & India \\
    \hline
    Answer of GraphRAG & Alsa Mall and Spencer Plaza are both located in Chennai, India [Data: India and Chennai Community (2391); Entities (4901, 4904); Relationships (9479, 1687, 5215, 5217)]. \\
    \hline
\end{tabular}
\label{tab:graphrag_example}
\end{table}

\subsection{Evaluation on Legal Benchmarks}  \label{subsec:legal_evaluation}

In this subsection, we present the performance of our approach on two legal benchmarks: LawBench~\cite{fei2023lawbench} and Open Australian Legal QA~\cite{butler-2023-open-australian-legal-dataset}. Before doing so, we provide a brief description of each benchmark.

\paragraph{LawBench} LawBench is a comprehensive legal benchmark for Chinese laws. It comprises 20 meticulously designed tasks aimed at accurately assessing the legal capabilities of LLMs. Unlike some existing benchmarks that rely solely on multiple-choice questions, LawBench includes a variety of task types that are closely related to real-world applications. These tasks encompass legal entity recognition, reading comprehension, crime amount calculation, and legal consulting, among others. Since not all tasks are RAG-oriented (e.g., reading comprehension), we have selected 6 specific tasks, which are detailed in Table~\ref{tab:lawbench_task}. The number of questions of each task is 500. 

\begin{table}[t]
\fontsize{9}{11}\selectfont
\renewcommand{\arraystretch}{1.1}
    \caption{Overview of LawBench tasks}
    \vspace{3mm}
    \centering
    {\footnotesize
    \begin{tabular}{c|c|c|c}
    \hline
         Task No. & Task & Type & Metric \\
         \hline
         1-1 & Statute Recitation  & Generation & F1\\
         \hline
         1-2 & Legal Knowledge Q\&A & Single Choice & EM \\
         \hline
         3-1 & Statute Prediction (Fact-based) & Multiple Choices & EM \\
         \hline
         3-2 & Statute Prediction (Scenario-based) & Generation & F1 \\
         \hline
         3-6 & Case Analysis & Single Choice & EM\\
         \hline
         3-8 & Consultation & Generation & F1\\
         \hline
    \end{tabular}}

    \label{tab:lawbench_task}
\end{table}

We also provide example questions of these tasks for the readers reference.
\begin{lstlisting}[language={}]
1-1: Answer the following question by directly providing the content of the article:What is the content of Article 76 of the Securities Law?
1-2: According to the 'Securities Law', which of the following statements about stock exchanges is incorrect? A: Without the permission of the stock exchange, no entity or individual may publish real-time securities trading information; B: The stock exchange may restrict trading on securities accounts that exhibit major abnormal trading conditions as needed, and report to the securities regulatory authority under the State Council for record; C: The accumulated property of a member-based stock exchange belongs to the members, and their rights are jointly enjoyed by the members; during its existence, the accumulated property may not be distributed to the members; D: The stock exchange formulates listing rules, trading rules, member management rules, and other relevant rules in accordance with securities laws and administrative regulations, and reports to the securities regulatory authority under the State Council for record.
3-1: Based on the following facts and charges, provide the relevant articles of the Criminal Law. Facts: The Yushu City, Jilin Province, accused that on November 15, 2015, the defendant He signed a car rental agreement with Guo, the owner of a taxi with license plate number xxx. The agreement stipulated a monthly rent of RMB 3,900.00, payable monthly. On January 19, 2016, without the knowledge of Guo, the defendant He concealed the truth and falsely claimed to be the owner of the taxi. He signed a car rental agreement with the victim Ma, with a monthly rent of RMB 3,800.00 and a rental period of one year, collecting a total of RMB 50,600.00 from Ma for one year's rent and vehicle deposit. On February 26, 2016, the taxi was retrieved by its owner Guo from the victim Ma. The victim Ma repeatedly asked the defendant He to return the rent and deposit, but the defendant He refused to return them. The prosecution provided evidence including the defendant's confession, the victim's statement, witness testimonies, and documentary evidence, and believed that the defendant He, with the purpose of illegal possession, defrauded others of their property by fabricating facts and concealing the truth during the signing and performance of the contract. The amount was relatively large, and his actions violated the provisions of Article xx of the Criminal Law of the People's Republic of China, and he should be held criminally responsible for xx. Charge: Contract Fraud. 
3-2: Please provide the legal basis according to the specific scenario and question, only the content of the specific legal provision is needed, each scenario involves only one legal provision. Scenario: A cargo ship arrives at the port of discharge, but the consignee fails to arrive in time to collect the goods. Under which legal provision can the captain unload the goods at another appropriate place?
3-6: One year after the bar opened, the business environment changed drastically, and all partners held a meeting to discuss countermeasures. According to the 'Partnership Enterprise Law,' the following voting matters are considered valid votes: A: Zhang believes that the name 'Tongcheng' is not attractive and proposes to change it to 'Tongsheng Bar.' Wang and Zhao agree, but Li opposes; B: In view of the sluggish business, Wang proposes to suspend operations for one month for renovation and reorganization. Zhang and Zhao agree, but Li opposes; C: Due to the urgent needs of the bar, Zhao proposes to sell a batch of coffee machines to the bar. Zhang and Wang agree, but Li opposes; D: Given the four partners' lack of experience in bar management, Li proposes to appoint his friend Wang as the managing partner. Zhang and Wang agree, but Zhao opposes.
3-8: Resident A rented out the house to B. With A's consent, B renovated the rented house and sublet it to C. C unilaterally altered the load-bearing structure of the house. Why can A request B to bear liability for breach of contract?
\end{lstlisting}

\paragraph{Open Australian Legal QA} The benchmark consists of 2,124 questions and answers synthesized by GPT-4 from the Australian legal corpus. All questions are of the generation type. One example is: "What is the landlord's general obligation under section 63 of the Act in the case of Anderson v Armitage [2014] NSWCATCD 157 in New South Wales?"

Evaluation results are listed in Table~\ref{tab:law_evaluation_results}, where we only compare to "GraphRAG Local", as it generally performs better than "GraphRAG Global" on these tasks.

\begin{table}[t]
\fontsize{9}{11}\selectfont
\renewcommand{\arraystretch}{1.1}
    \caption{Evaluation Results on Legal Benchmarks (Metric is \textbf{F1 / EM} as indicated in Table~\ref{tab:lawbench_task})}
    \vspace{3mm}
    \centering
    \begin{tabular}{c|c|c|c|c}
    \hline
         \multicolumn{2}{c|}{Task} & Zero-Shot CoT & GraphRAG Local & Ours (N=5) \\
         \hline
         \multirow{6}{*}{LawBench}& 1-1 & 21.31 &	\underline{23.27}	& \textbf{78.58}\\
         & 1-2 &  54.24	& \underline{62.60}	& \textbf{70.60} \\
         & 3-1 & 53.32	& \underline{74.60} & \textbf{83.16}\\
         & 3-2 & \underline{27.51}	& 25.98	& \textbf{46.05}  \\
         & 3-6 &  \underline{51.16} &	47.64 &	\textbf{61.91}  \\
         & 3-8 &  17.44	& \underline{18.43}	& \textbf{23.58} \\
         \hline
         \multicolumn{2}{c|}{Open Australian Legal QA} & 25.10 & \underline{34.35} &  \textbf{63.34}\\
         \hline
    \end{tabular}

    \label{tab:law_evaluation_results}
\end{table}

\begin{table}[t]
\fontsize{9}{11}\selectfont
\renewcommand{\arraystretch}{1.1}
    \caption{Evaluation Results on Legal Benchmarks (Metric is \textbf{Acc})}
    \vspace{3mm}
    \centering
    \begin{tabular}{c|c|c|c|c}
    \hline
         \multicolumn{2}{c|}{Task} & Zero-Shot CoT & GraphRAG Local & Ours (N=5) \\
         \hline
         \multirow{6}{*}{LawBench}& 1-1 & 1.23 & \underline{16.60} & \textbf{90.12} \\
            & 1-2 & 54.00 & \underline{63.40} & \textbf{70.60} \\
            & 3-1 & 49.90 & \underline{75.40} & \textbf{88.82} \\
            & 3-2 & 15.83 & \underline{27.60} & \textbf{67.54} \\
            & 3-6 & 51.12 & \underline{57.00} & \textbf{62.73} \\
            & 3-8 & 49.70 & \underline{58.80} & \textbf{61.72} \\
         \hline
         \multicolumn{2}{c|}{Open Australian Legal QA} & 16.48 & \underline{88.27} & \textbf{98.59} \\
         \hline
    \end{tabular}

    \label{tab:law_evaluation_results_acc}
\end{table}

For the aforementioned reasons, we also use GPT-4 to evaluate all experimental results, reporting the accuracy (\textbf{Acc}) in Table~\ref{tab:law_evaluation_results_acc}. When comparing the results in Table~\ref{tab:law_evaluation_results} and Table~\ref{tab:law_evaluation_results_acc}, we observe that the order of the results is preserved, even though some metrics change significantly. In the following section, we aim to identify the reasons behind these changes, which may provide valuable insights for designing better metrics to evaluate RAG frameworks in the future.
\begin{enumerate}
    \item The accuracy of our approach increases significantly for generation tasks (1-1, 3-2, Open Australian Legal QA). For these tasks, our answers are often semantically equivalent but syntactically different from the golden answers. This explains the improved metric performance, as GPT-4 can compare the semantic content of the answers. This also applies to the "GraphRAG Local" results for the "Open Australian Legal QA" task.
    \item The accuracy of "GraphRAG Local" decreases for generation tasks 1-1 and 3-2. These tasks involve statute recitation and prediction, requiring the retrieval of specific articles. Upon detailed examination, We find that "GraphRAG Local" often fails to retrieve the correct articles or references the wrong ones, but it tends to repeat the legal information. Therefore, token-level recall can be improved by simply rephrasing legal names and common prefixes, such as "According to XX law, XX articles...".
    \item Both our approach and "GraphRAG Local" show significant accuracy improvements on task 3-8. Besides the reason mentioned in the first point, the quality of the golden answers may also contribute to this difference. The questions and golden answers in task 3-8 are sourced from a consulting website, resulting in varying quality. For example, one question asks "Do the children from the original marriage have an obligation to support the father?" However, the provided golden answer includes an irrelevant article, "Article 1067," which pertains to parents' obligations to support minor children.
\begin{lstlisting}[language={}]
Question: In the case where both parents are divorced and have formed their own families with new children, and according to the court's judgment, the father is required to pay monthly child support to the mother until the child is 18 years old. Do the children from the original marriage have an obligation to support the father?
Reference Answer: In our country, biological children have an obligation to support their parents who are divorced. The relationship between children and parents does not dissolve because of the divorce of the children or parents. Supporting parents is a legal obligation of the children. If the children do not support the elderly, the parents can directly file a lawsuit in the People's Court requesting the court to order the daughter to pay support. Legal basis: Article 1067 of the Civil Code of the People's Republic of China states that if parents do not fulfill their obligation to support, minor children or adult children who cannot live independently have the right to request support from their parents. If adult children do not fulfill their obligation to support, parents who are incapacitated or have difficulties in living have the right to request support from adult children. Article 1084 states that the relationship between parents and children does not dissolve because of the parents' divorce. After the divorce, whether the children are directly raised by the father or the mother, they are still the children of both parents. Lawyer's explanation: After the parents are divorced, adult children still have the obligation to support both parents. The obligation of adult children to support their parents does not change because of changes in the parents' relationship. If adult children do not fulfill their obligation to support, parents who are incapacitated or have difficulties in living have the right to request support from adult children.
Our Answer: Yes, there is an obligation to support. Legal basis: According to Article 1069 of the Civil Code of the People's Republic of China, it is stipulated that: 'The obligation of children to support their parents does not terminate because of changes in the parents' marital relationship.' Therefore, even if the parents are divorced and remarried with new children, the original children still have an obligation to support their parents.
\end{lstlisting}
    \item The accuracy of all methods on choice tasks 1-2, 3-1, and 3-6 almost coincides with the F1 score, as expected. An exception is task 3-1, where the difference is mainly due to GPT-4's capacity to understand Chinese, particularly in distinguishing numbers in Arabic and Chinese. In Chinese law, all numbers are written in Chinese, while in the golden answers, all numbers are given in Arabic.
\end{enumerate}

\begin{figure}[ht]
	\begin{center}
		\includegraphics[width=0.99\linewidth]{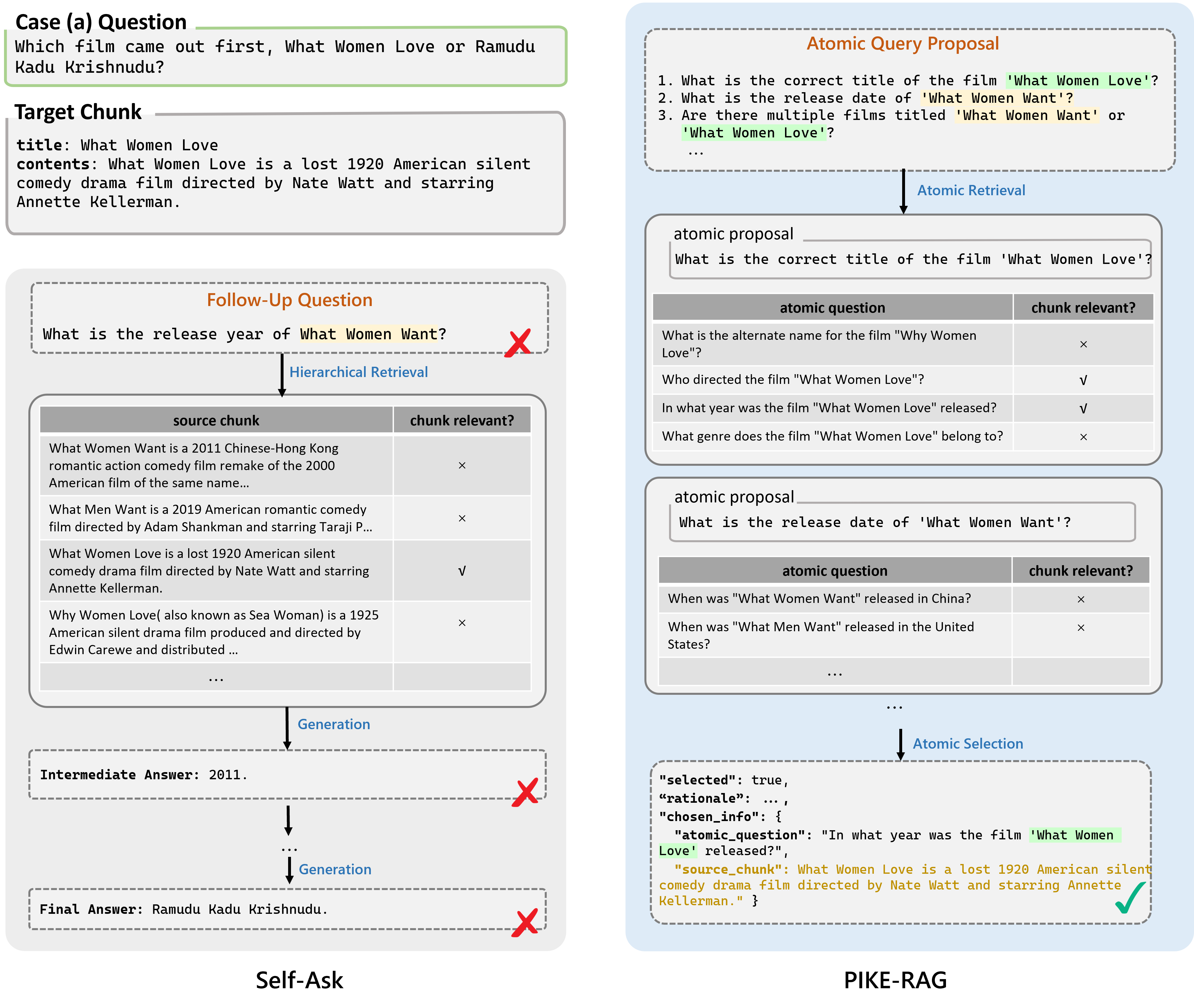}
	\end{center}
    \caption{Case (a): Given the lesser-known film "What Women Love" as opposed to the more popular "What Women Want," single-path methods like Self-Ask on the left are predisposed to generating follow-up questions about the latter, leading to an incorrect final answer. Conversely, PIKE-RAG can effectively discern the intended meaning of the original question by positing several atomic queries and postpone the task understanding to atomic selection phase with relevant atomic questions provided, and subsequently arriving at an accurate conclusion.} 
    \label{fig:case_a}
    \vspace{-2mm}
\end{figure}

\subsection{Experiments on Domain Aligned Atomic Proposers}

Given that all methods demonstrate notably low accuracy scores on the MuSiQue dataset, we specifically selected this dataset to investigate whether implementing a fine-tuned atomic proposer could potentially enhance overall performance metrics and improve the quality of results. 

\paragraph{Data Collection} To establish a comprehensive and efficient experimental framework, we initiate our process by randomly sampling 1000 questions from the dataset. These questions are subsequently divided into two distinct sets following a standard split ratio: 80\% of the questions are designated as training data, specifically utilized for collecting and analyzing decomposition trajectories, while the remaining 20\% are reserved as an evaluation dataset to assess the model's performance. For each question within the training dataset, we meticulously follow the procedural framework outlined in Figure~\ref{fig: data_collection}, systematically generating follow-up questions in a stepwise progression until we successfully obtain a decomposition trajectory that leads to the correct answer. We establish a maximum limit of 10 rounds per question. The results of our data collection process prove to be efficient - we successfully obtain correct answers along with their corresponding decomposition trajectories for 84\% of the training data, which represents a substantial improvement compared to the baseline accuracy of 58\% achieved without implementing context sampling techniques.

\paragraph{Training Method} Our training methodology is designed to optimize model performance by focusing exclusively on decomposition trajectories that successfully lead to correct answers. The data structure for each record follows a format that captures the complete problem-solving path: $(q, [(q^1, a^1), \ldots, (q^t, a^t)], a)$, where $q$ and $a$ are the original question and its answer, respectively. Each pair $(q^i, a^i)$ in the trajectory is composed of an atomic question and its answer. This structured approach ensures that we maintain a clear record of the decomposition process from start to finish.

\begin{algorithm}[ht]
\caption{Transform each decomposition trajectory into data pairs for SFT}
\label{alg:decompose-sft}
\begin{algorithmic}[1]
    \State Let $(q, [(q^1, a^1), \ldots, (q^t, a^t)], a)$ be the given decomposition trajectory.
    \State Initialize an empty list to store pairs of prompts and responses $\mathcal{D}_{\mathit{sft}}=\emptyset$.
    \State \# Prepare training data when a further decomposition is necessary.
    \For {$i=1,2,\ldots,t$}
        \State \# $\mathit{prompt}_x$ takes the original question plus any previously collected sub-questions and answers as input.
        \State $x_i=\mathit{prompt}_x(q, [(q^1, a^1), \ldots, (q^{i-1}, a^{i-1})])$
        \State \# $\mathit{prompt}_y$ takes two inputs: a decomposition indicator and an optional sub-question.
        \State $y_i=\mathit{prompt}_y(True, a^i)$
        \State $\mathcal{D}_{\mathit{sft}}\xleftarrow{}(x_i, y_i)$
    \EndFor
    \State \# Prepare training data when no further decomposition is necessary.
    \State $x_{t+1}=\mathit{prompt}_x(q, [(q^1, a^1), \ldots, (q^t, a^t)])$
    \State $y_{t+1}=\mathit{prompt}_y(False, None)$
    \State $\mathcal{D}_{\mathit{sft}}\xleftarrow{}(x_{t+1}, y_{t+1})$
\end{algorithmic}
\end{algorithm}

For our LLM training, we employ supervised fine-tuning (SFT) to enhance model performance. Our process converts each trajectory into structured training data pairs of prompts and responses. These pairs are formatted as $(x_i, y_i)$ according to Algorithm~\ref{alg:decompose-sft}. The algorithm incorporates two key functions: $\mathit{prompt}_x$ and $\mathit{prompt}_y$, which generate contextual descriptions based on input parameters. These functions are integral to the transformation process, and we detail their implementation in the following sections.

\begin{minipage}{\linewidth}
\begin{dialogbox}[Prompt Template for $\mathit{prompt}_x$]
\footnotesize
Based on the given information, determine whether a follow-up question is necessary or not. \\
**Original Question**\\
\textcolor{darkbrown}{\{the original problem\}}\\
**Existing Context**\\
\textcolor{darkbrown}{\{A list of sub-questions and their answers\}}\\
Make sure your output align with the following format:\\
<decompose>False</decompose>\\
OR\\
<decompose>True</decompose>\\
<sub-question>a follow-up question</sub-question>\\
\end{dialogbox}
\end{minipage}

\begin{minipage}{\linewidth}
\begin{dialogbox}[Prompt Template for $\mathit{prompt}_y$ if a sub-question is provided]
\footnotesize
<decompose>True</decompose>\\
<sub-question>\textcolor{darkbrown}{\{The given sub-question\}}</sub-question>
\end{dialogbox}
\end{minipage}

\begin{minipage}{\linewidth}
\begin{dialogbox}[Prompt Template for $\mathit{prompt}_y$ if no sub-question is provided]
\footnotesize
<decompose>False</decompose>
\end{dialogbox}
\end{minipage}

This transformation is particularly efficient because it generates multiple training instances from each trajectory - specifically, for a trajectory containing $t$ sub-questions, we generate $t+1$ distinct training data points, maximizing the learning potential from each successful decomposition.

For all experiments, we use a learning rate of $1.5e^{-5}$ and parameter-efficient fine-tuning (PEFT) with LoRA configuration parameters (lora=16, alpha=64). To ensure reproducibility, we maintain all other hyperparameters at their default values as specified in reference~\cite{vonwerra2022trl}. We run all experiments three times with different random seeds and report the mean results. We perform the experiments on a computing node equipped with a single NVIDIA A100-80G GPU for all training processes.

\paragraph{Experiment Results} Table~\ref{tab:decompose-ft-exp} presents our experimental findings. We select three open-source LLMs with varying parameter sizes and computational capabilities as our base models for atomic proposers: meta-llama/Llama-3.1-8B~\cite{llama3modelcard}, Qwen/Qwen2.5-14B~\cite{qwen2}, and microsoft/phi-4~\cite{abdin2024phi4technicalreport}. Through SFT, we develop these into fine-tuned atomic proposers. For inference, we integrated two advanced LLMs-GPT-4o (version 2024-11-20) and Llama-3.1-70B-Instruct~\cite{llama3modelcard}-to process the retrieved data and generate the final answers. These models were chosen for their enhanced capabilities in delivering accurate, high-quality outputs.

For illustration, the second row and second column in Table~\ref{tab:decompose-ft-exp} shows our use of the original meta-llama/Llama-3.1-8B as the atomic proposer for question decomposition. After gathering sufficient context, we use GPT-4o to generate the final answer. In contrast, the second row and third column shows results when using a fine-tuned version of meta-llama/Llama-3.1-8B as the atomic proposer, while still using GPT-4o for final answer generation. The difference in values demonstrates how fine-tuning atomic proposers with domain-specific decomposition trajectories improves performance.

\begin{table}[t]
\fontsize{9}{11}\selectfont
\renewcommand{\arraystretch}{1.1}
    \centering
    \caption{Evaluation Effects of Domain Aligned Atomic Proposers (Columns represent different generation LLMs while rows denote smaller LMs that we adopt as atomic decomposer.)}
    \vspace{2mm}
    \begin{tabular}{c|c|c|c|c}
        \hline
         & GPT-4o & GPT-4o+FT & Llama-3.1-70B-Instruct & Llama-3.1-70B-Instruct+FT\\
        \hline
        Llama-3.1-8B & 47.83\% & 62.14\% & 48.37\% & 58.70\% \\
        \hline
        Qwen2.5-14B & 56.52\% & 63.95\% & 57.61\% & 63.04\% \\
        \hline
        phi-4-14B  & 60.33\% & 65.76\% & 58.70\% & 62.50\% \\
        \hline
    \end{tabular}
    \label{tab:decompose-ft-exp}
\end{table}

\subsection{Real Case Studies}  \label{subsec:real_cases}

This section presents three case studies from our evaluation benchmark to illustrate the underlying principles of our proposed decomposition pipeline, as detailed in Algorithm~\ref{alg:ka-decompose}. Through these real-world examples, we aim to highlight the benefits of our systematic approach. These cases will shed light on how each step of the pipeline contributes to improved performance and the insights gained from their implementation.

Our task decomposition strategy involves generating multiple atomic queries rather than producing a single deterministic follow-up question, as demonstrated in the Self-Ask approach. Contemporary decomposition methods typically employ a generative model to formulate a singular follow-up question. However, this approach carries an intrinsic risk of generating erroneous questions, potentially leading to an incorrect decomposition pathway and, ultimately, an erroneous answer. Consider the Case (a) depicted in Figure~\ref{fig:case_a}, where the original question pertains to a film titled "What Women Love." Due to the existence of a more prominent film, "What Women Want," the employed language model tends to `correct' the original question. Consequently, methods like Self-Ask (as shown on the left side of Figure~\ref{fig:case_a}) generate only one follow-up question related to this erroneously assumed object. In the illustrated instance, although the target chunk has been retrieved due to the similarity in embeddings, a `false' intermediate answer is produced for the `false' follow-up question, culminating in an incorrect final response. In contrast, our methodology posits atomic queries concerning both "What Women Love" and "What Women Want," thereby seeking to clarify the true intent of the initial question. With both films existing and relevant atomic questions being retrieved, our approach subsequently gains the advantage of verifying the question's intent and selecting the correct and most pertinent chunk during the atomic selection phase.

\begin{figure}[ht]
	\begin{center}
		\includegraphics[width=0.95\linewidth]{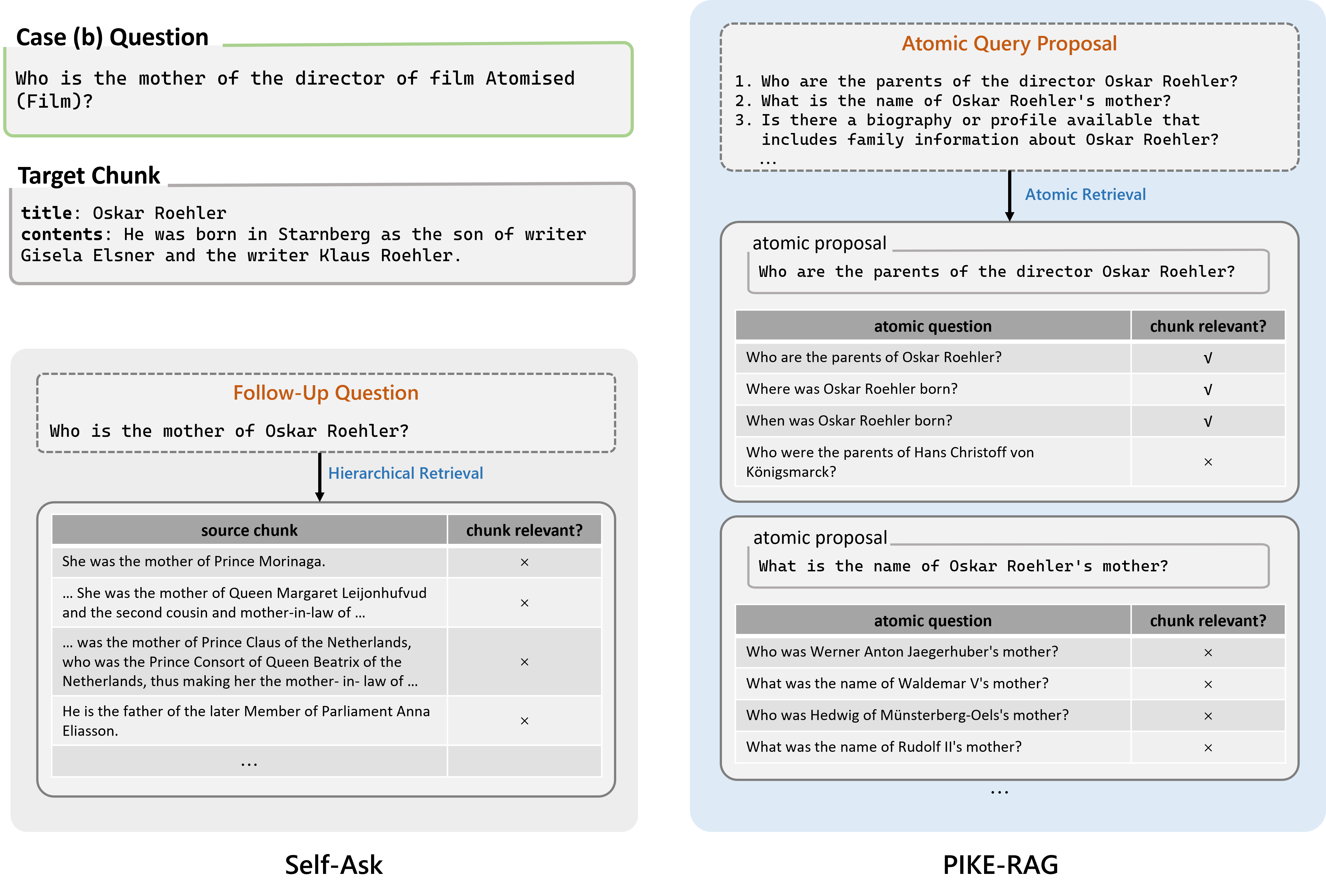}
	\end{center}
	\caption{Case (b): By proposing multiple atomic queries, PIKE-RAG effectively retrieves the relevant knowledge chunk, whereas the single deterministic follow-up question approach employed by Self-Ask fails to align with the knowledge base's schema, resulting in a retrieval failure.} 
    \label{fig:case_b}
    \vspace{-2mm}
\end{figure}

Furthermore, the discrepancy between the formulation of the corpus and the query, is another critical factor advocating for a multi-query approach over a singular deterministic one. The presentation gap can impede the retrieval process even when the generated follow-up question is semantically accurate. For instance, as illustrated in Case (b) in Figure~\ref{fig:case_b}, a single-path method such as Self-Ask on the left side might directly inquire `Who is the mother of Oskar Roehler?' However, the knowledge base articulates familial relationships using a different schema, `A is the son of B and C' in this case, thus the retrieval process falters despite the correctness of the question. Even when we applied the hierarchical retrieval to Self-Ask, the Self-Ask with Hierarchical Retrieval did not succeed in bridging this gap. In contrast, our approach, which generates multiple atomic queries, encompasses a broader range of phrasings that correspond to the diverse representations in the knowledge base. In the depicted case, while the atomic query specifically asking for Oskar Roehler's mother encounters the same retrieval issue, an alternative query seeking information about his parents successfully retrieves the target chunk. This exemplifies how our method's flexibility in query generation enhances the likelihood of aligning with the knowledge base's structure and obtaining accurate information.

Our methodology emphasizes the retrieval of atomic questions rather than directly retrieving chunks. This design choice is exemplified in Case (b) depicted in Figure~\ref{fig:case_b}. The knowledge chunk in the corpus is structured using the pattern `A ... as the son of B and C', which poses challenges for direct retrieval by queries such as `Who is the mother of ...'. In our specialized knowledge base, such direct queries tend to retrieve chunks conforming to the patterns `A is the mother of B' or `A is the father of B'. By utilizing atomic questions as intermediaries for retrieval, our approach effectively narrows the gap between a single query and the multiple sentence structures found in the knowledge base. It facilitates bridging the expression pattern differences exemplified by `the mother of' versus `the son of' in this scenario.

\begin{figure}[t]
	\begin{center}
		\includegraphics[width=0.99\linewidth]{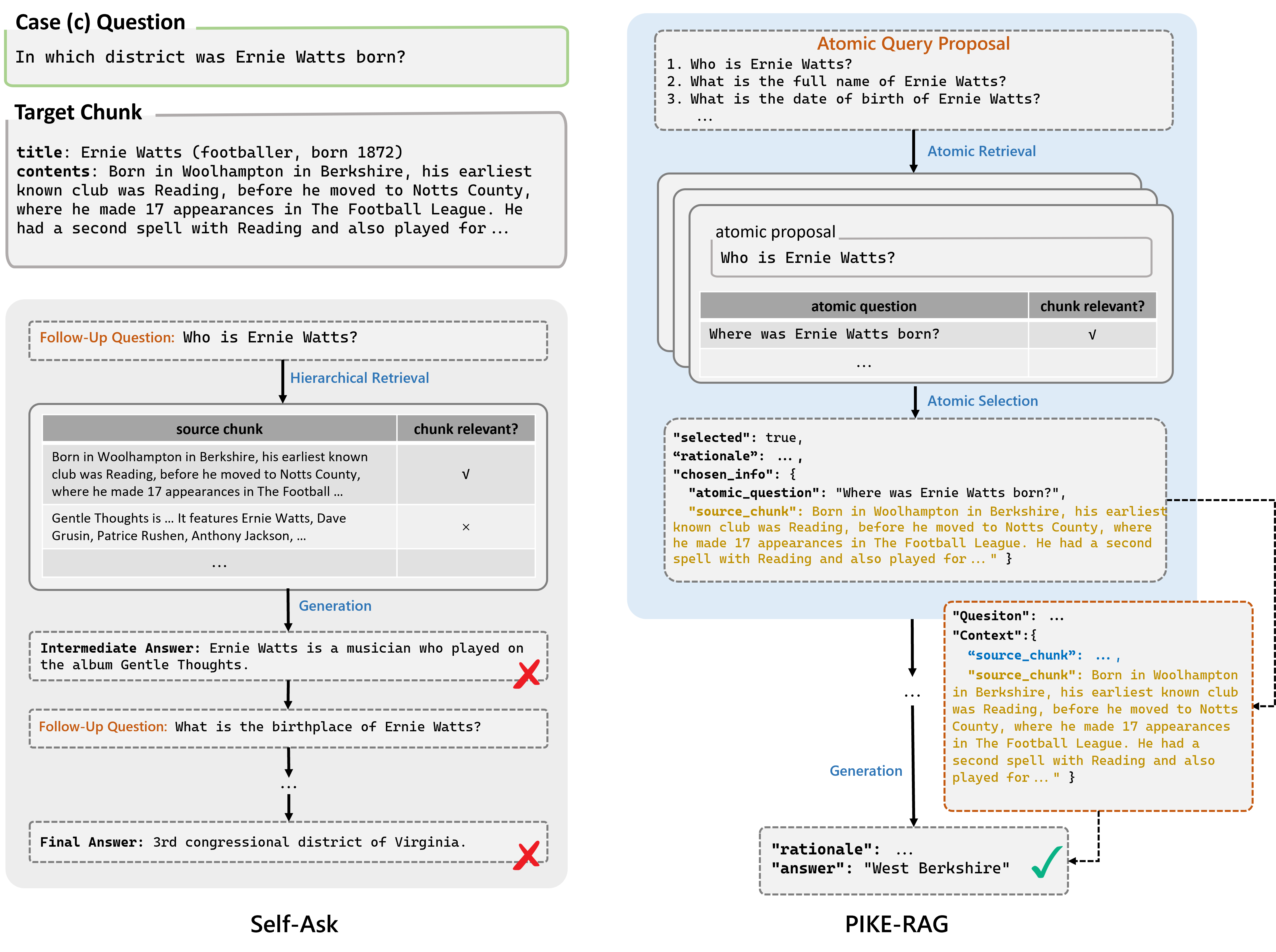}
	\end{center}
	\caption{Case (c): PIKE-RAG's benefits from leveraging a concise list of atomic questions for targeted selection and retaining full chunks for rich contextual support. Conversely, Self-Ask's approach, although successful in retrieving relevant chunks, is compromised by its dependency on intermediate answers for context, which ultimately results in the generation of incorrect final answers.} 
    \label{fig:case_c}
\end{figure}

In contrast to methods like Self-Ask, which only retains intermediate answers for subsequent processing, our method preserves the entire chunk as contextual information. During the atomic selection phase, we present a list of atomic questions as candidate summaries of the relevant content from the original chunk. This strategy significantly reduces token usage and simplifies the process of selecting the pertinent information. Case (c) in Figure~\ref{fig:case_c} demonstrates the dual benefits of our approach: first, by selecting from a curated list of atomic questions, we streamline the identification of relevant information; second, by retaining the entire selected chunk rather than just the intermediate answer, we ensure a rich context is maintained for accurate and comprehensive subsequent processing. While the Self-Ask method on the left retrieves the target chunk, it fails to correctly identify the pertinent `Ernie Watts' due to the excessive contextual information. Since retrieved chunks in Self-Ask are discarded after generating an intermediate answer, the method potentially follows an incorrect pathway, leading to an inaccurate conclusion. In contrast, our approach can efficiently filter and select the appropriate atomic question from a concise list. Although the atomic question in this round pertains to the role of Ernie Watts, there is no need to inquire further about his birthplace, as this information is encapsulated within the selected chunk, which remains available for context in subsequent rounds.

\section{Conclusion}
To address the diverse challenges faced by RAG systems in industrial applications, we propose that the core foundation of RAG systems should extend beyond traditional retrieval mechanisms to the effective construction and utilization of specialized knowledge and rationale.
Therefore, we introduce a new paradigm that classifies tasks based on their difficulty in knowledge extraction, comprehension, and utilization, providing a novel framework for system design and evaluation. Applying this paradigm allows for phased exploration of RAG capabilities, which facilitates the progressive refinement of RAG algorithms and the staged implementation of RAG applications.
Moreover, we introduce the specialized Knowledge and Rationale Augmented Generation (PIKE-RAG) framework, focusing on specialized knowledge extraction and rationale construction. PIKE-RAG effectively extracts, comprehends, and organizes specialized knowledge and construct coherent rationale for accurate answers, offering customizable system capabilities to meet varying requirements. Additionally, we propose knowledge atomizing and knowledge-aware task decomposition to tackle complex questions, such as multihop queries, achieving significant performance improvements on various open-domain and legal benchmarks. 

\clearpage
\bibliographystyle{plain}
{\small
  \bibliography{ref}
}











\end{document}